%% file: main.tex
\documentclass{article}

\PassOptionsToPackage{numbers, compress}{natbib}

\usepackage[preprint]{main}

\usepackage[utf8]{inputenc}
\usepackage[T1]{fontenc}
\usepackage{hyperref}
\usepackage{url}
\usepackage{booktabs}
\usepackage{amsfonts}
\usepackage{nicefrac}
\usepackage{microtype}
\usepackage{xcolor}

\usepackage{orcidlink}

\usepackage{amsmath}
\usepackage{amssymb}
\usepackage{bm}
\usepackage{bbm}
\usepackage{algorithm}
\usepackage{multirow}
\usepackage{graphicx}
\usepackage{xspace}

\newcommand{\ie}{\textit{i.e.}\xspace}
\newcommand{\eg}{\textit{e.g.}\xspace}

\usepackage{tikz}
\usetikzlibrary{positioning,arrows.meta,fit,calc,backgrounds}
\usepackage{float}
\usepackage{algpseudocode}

\author{
Rui Song$^{1,2,}$\thanks{Equal contribution.}\quad
Tianhui Cai$^{1,*}$\quad
Markus Gross$^{3}$\quad
Xingcheng Zhou$^{3}$\quad
Zewei Zhou$^{1}$
\\
\textbf{Zhiyu Huang$^{1}$\quad
Olaf Wysocki$^{2}$\quad
Jiaqi Ma$^{1}$}
\\[0.10cm]
{\small $^{1}$University of California, Los Angeles \quad
$^{2}$University of Cambridge \quad
$^{3}$Technical University of Munich}
}

\begin{document}

\newcommand{\myMethod}{\mbox{ConFixGS}\xspace}

\title{ConFixGS: Learning to Fix Feedforward 3D Gaussian Splatting with Confidence-Aware Diffusion Priors in Driving Scenes}

\maketitle
\begingroup
\renewcommand{\thefootnote}{}
\footnotetext{Corresponding author. Email: \texttt{rruisong@ucla.edu}.}
\endgroup

\begin{figure}[tbh]
   \centering
   \includegraphics[width=\textwidth]{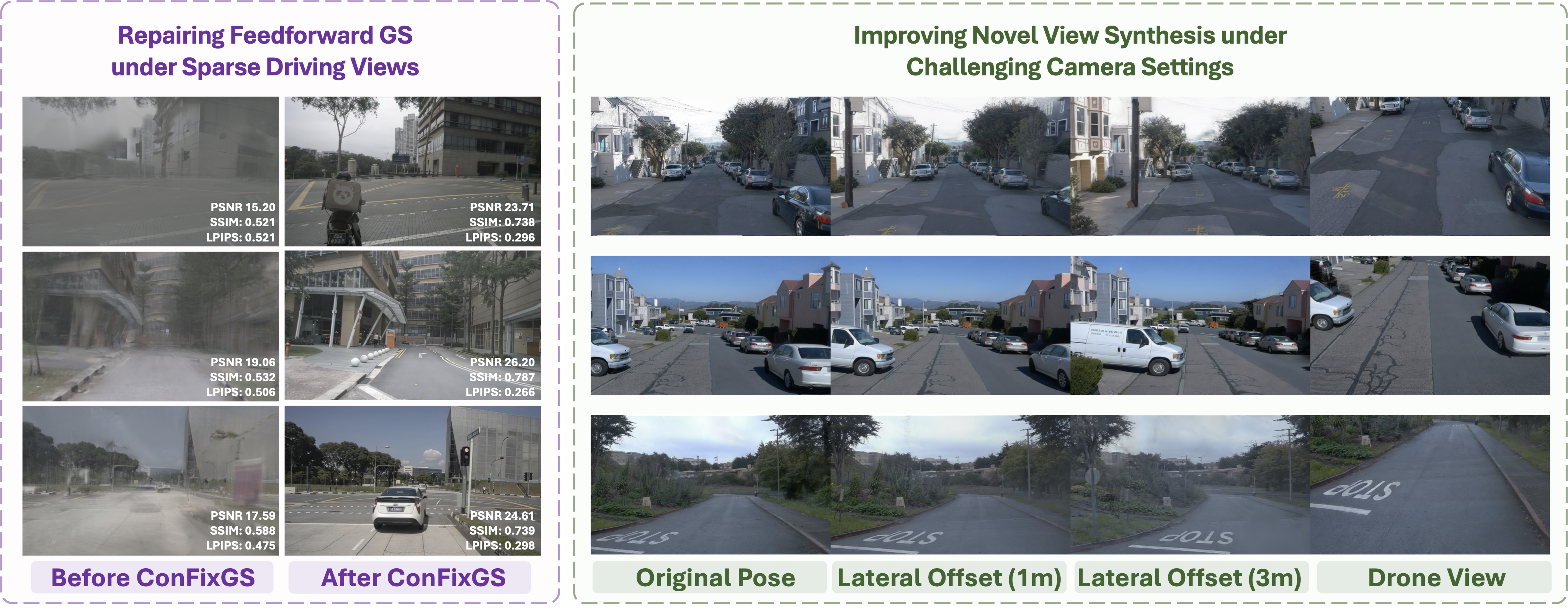}
   \caption{\textbf{\myMethod{}: a plug-and-play repair of feedforward 3DGS in sparse driving scenes.}
\emph{Left:} Our confidence-guided method enhances state-of-the-art feedforward backbones, yielding better novel view rendering and consistent gains across metrics.
\emph{Right:} The repaired 3D Gaussians generalize well beyond the original camera trajectory, supporting novel view synthesis under large lateral offsets and elevated drone-like viewpoints settings where unrepaired feedforward 3DGS typically fails.
}
    \label{fig:teaser}
\end{figure}

\begin{abstract}

Feedforward 3D Gaussian Splatting (3DGS) often struggles in trajectory-based sparse-view driving scenes. Existing Gaussian repair methods mainly target optimization-based 3DGS, while diffusion-based repair is typically restricted to iterative refinement near observed viewpoints, leaving feedforward 3DGS repair underexplored.
We propose \textbf{ConFixGS}, a plug-and-play method that learns to fix feedforward 3DGS with confidence-aware diffusion priors. Starting from a pretrained feedforward model, ConFixGS generates diffusion-enhanced local pseudo-targets and validates them through reprojection-based cross-checking against support views. The resulting dense confidence maps guide refinement, enhancing reliable details while suppressing hallucinated or inconsistent evidence.
On Waymo, nuScenes, and KITTI, ConFixGS improves challenging novel view synthesis, with PSNR gains of up to 3.68 dB and FID reduced by nearly half. Our results highlight confidence-aware fusion of generative priors and support-view consistency as a key principle for robust feedforward 3D driving scene reconstruction.

\end{abstract}

\input{sections/01_intro}
\input{sections/02_related_work}
\input{sections/03_method}
\input{sections/05_exp}
\input{sections/06_conclusion}

\newpage

\bibliographystyle{unsrt}
\bibliography{main}

\newpage

\begin{center}
\LARGE \textbf{Technical Appendix}
\end{center}
\appendix

We provide additional implementation details, methodological analysis, efficiency evaluation, and qualitative results to support the main paper. To facilitate reproducibility, Sec.~\ref{sec:impl} details the complete benchmark configuration, including dataset protocols and all key hyperparameters used in our experiments. Sec.~\ref{sec:gs_grad} provides additional methodological details on our confidence-weighted objective and gradients, clarifying how the support-validated confidence is incorporated into optimization and densification. Sec.~\ref{sec:efficiency} reports a complexity and efficiency analysis. Finally, Sec.~\ref{sec:add_comp} presents additional qualitative results in three parts: Sec.~\ref{sec:nvs} novel view synthesis along out-of-distribution trajectories, including lateral offsets and a virtual drone view; Sec.~\ref{sec:further_backbone} plug-and-play enhancement results on additional feedforward backbones; and Sec.~\ref{sec:difix3D} further visual comparisons with Difix3D+. The videos complement these results by comparing our method against multiple baselines, highlighting the stability of ConFixGS under challenging viewpoint changes and its effectiveness in enhancing state-of-the-art feedforward Gaussian Splatting methods. 

\input{sections/07_imp_details}
\input{sections/08_opt_details}
\input{sections/10_efficiency}
\input{sections/09_add_visual}

\end{document}

%% file: sections/01_intro.tex
\section{Introduction}
\label{sec:intro}

Accurate 3D reconstruction of driving scenes is essential for autonomous driving simulation, novel view synthesis, and scenario replay~\cite{zhou2025hugsim,Cao2025CORL,zhao2026bridgesim}. While vehicle-mounted cameras provide scalable real-world observations, they are typically collected along forward-moving trajectories, leading to sparse, anisotropic views with limited overlap, wide baselines, and severe occlusions. 3D Gaussian Splatting (3DGS) enables real-time high-fidelity rendering~\cite{kerbl3Dgaussians}, but conventional 3DGS requires costly per-scene optimization and is difficult to scale. Feedforward 3DGS addresses this limitation by predicting Gaussian primitives from sparse input images in a single forward pass~\cite{charatan2023pixelsplat,chen2024mvsplat,wang2024freesplat,hong2025pfplat,ye2024no,wewer2024latentsplat,smart2024splatt3r,ye2025yonosplat}, making it increasingly appealing for driving-scene modeling~\cite{hou2026drivingscene,song2026energs,lin2025vgd,han2025ggs}. However, in trajectory-based sparse-view settings, feedforward models often produce blurred objects, low contrast, and
texture smearing.

Our key observation is that feedforward 3DGS exhibits different behavior at varying trajectory scales. When applied to a small set of nearby, highly overlapping views, it often produces sharper and more reliable local novel views than when it is forced to reconstruct a large scene from many sparsely connected trajectory images, and applying a 2D diffusion prior on top of these local renderings produces the repairable pseudo-targets (Fig.~\ref{fig:local_vs_global_fw} on Waymo, nuScenes, and KITTI). Although such local reconstructions are incomplete, they provide useful intermediate renderings that can serve as repair candidates for a global representation. This suggests a natural strategy: decompose a sparse driving trajectory into local overlapping subsets, generate local pseudo views, enhance them with strong 2D priors, and use them to refine the initial global feedforward reconstruction.

Recent diffusion-assisted 3DGS repair methods show that 2D generative priors can enhance degraded renderings and distill high-frequency details back into 3D representations~\cite{wu2025difix3d+,liu20243dgs,wang2025fixinggs,wei2025gsfix3d,yin2025gsfixer,paliwal2025ri3d,yu2024lm}. However, these methods are mostly designed for optimization-based 3DGS or rely on iterative repair near the original viewpoints, limiting their deployment in a feedforward setting. More critically, directly optimizing a global 3DGS from independently diffusion-enhanced local views is ill-posed: diffusion outputs are stochastic and not inherently multi-view consistent. Treating all pseudo views as equally reliable can introduce hallucinated textures, ambiguous geometry, ghosting, and spurious Gaussians. Thus, the central challenge is not whether diffusion priors are useful, but how to decide which generated details are confident enough to supervise feedforward 3DGS repair.

In this paper, we take a different view: \emph{diffusion-enhanced pseudo views should not be blindly distilled into 3D; their confidence should be explicitly estimated through the geometry of the feedforward reconstruction itself.} Inspired by reprojection-based consistency reasoning in controllable novel view and video synthesis~\cite{yan2025streetcrafter,seo2024genwarp,muller2024multidiff,yang2024driving,fan2025freesim,wang2024freevs}, we propose \textbf{\myMethod}, a confidence-guided approach for repairing feedforward 3DGS in sparse driving scenes. Starting from an initial feedforward reconstruction, \myMethod builds local feedforward reconstructions from overlapping view subsets and enhances their rendered novel views with a 2D diffusion prior. Each enhanced view is treated as a candidate pseudo-target rather than direct ground truth. To estimate its reliability, we unproject pseudo-target pixels using depth rendered from the initial reconstruction, reproject them into nearby real support views, and compare the reprojected appearance with observed images. This step produces a dense confidence map that identifies which generated details are supported by multi-view evidence and which regions are uncertain or likely hallucinated.

The resulting confidence maps guide a global 3DGS repair initialized from the original feedforward prediction: reliable diffusion details are injected into the Gaussian representation, while uncertain regions are down-weighted to avoid corrupting geometry or spawning spurious Gaussians. Since \myMethod repairs rather than reconstructs from scratch, it requires fewer than 1K optimization iterations, compared with the 30K--50K iterations commonly used by optimization-based driving-scene 3DGS pipelines. Experiments on Waymo, nuScenes, and KITTI show consistent improvements over strong feedforward baselines, and extend to challenging novel view settings with large lateral camera offsets and elevated drone-like viewpoints, as shown in Fig.~\ref{fig:teaser}. Our contributions are fourfold:

\begin{itemize}
    \item We identify a scale-induced failure mode of feedforward 3DGS in driving scenes: with long observation sequences, global processing can oversmooth local details, whereas processing nearby subsets with the same backbone preserves sharper and more repairable evidence.

    \item We propose \textbf{\myMethod}, the first plug-and-play repair approach for feedforward 3DGS without backbone retraining, reducing artifacts, and yielding high-quality novel views.

    \item We estimate continuous $[0,1]$ confidence maps for diffusion-enhanced local pseudo-targets via support-view reprojection, separating reliable generative details from hallucinations.

    \item We use these confidence maps to guide global Gaussian repair, modulating both photometric supervision and densification to improve large-offset and elevated-view rendering.
\end{itemize}

%% file: sections/02_related_work.tex
\section{Related Work}
\label{sec:related}

\noindent\textbf{3D Reconstruction for Driving Scenes.}
Driving-scene reconstruction is central to autonomous driving, supporting realistic simulation, controllable data generation, and evaluation under rare trajectories and viewpoints. Although neural radiance fields (NeRFs) and their unbounded-scene variants established high-quality view synthesis as a reconstruction objective~\cite{mildenhall2021nerf,barron2021mip,zhang2020nerf++,barron2022mip}, 3DGS represents scenes with explicit anisotropic Gaussian primitives and enables real-time rendering via efficient rasterization~\cite{kerbl3Dgaussians}. Follow-up 3DGS methods improve anti-aliasing, geometry, surfaces, propagation, resource usage, and sparse-view robustness~\cite{yu2024mip,lu2024scaffold,Huang2DGS2024,chen2024pgsr,guedon2024sugar,cheng2024gaussianpro,mallick2024taming,li2024geogaussian,turkulainen2025dn,chung2024depth,li2024dngaussian,huang2025fatesgs,zhang2025cdgs,huang2025det,fang2025nerf}. For outdoor and autonomous-driving scenes, prior work further leverages street-view assumptions, dynamic decomposition, laser scanning or depth priors, object structure, and large-scale spatio-temporal modeling~\cite{guo2023streetsurf,zhou2024drivinggaussian,yan2024street,jiang2024li,shen2025lidar,lee2025geomgs,xia2025d,cui2025streetsurfgs,miao2025evolsplat,hess2025splatad,wu20244d,xu2025ad,chen2026periodic,yang2024storm}. However, driving data is inherently trajectory-based, featuring sparse forward-moving viewpoints, wide baselines, limited overlap, and large unobserved regions. These properties make per-scene optimization slow and fragile, especially for novel views far from the original vehicle trajectory.

\noindent\textbf{Feedforward 3DGS.}
To avoid costly per-scene optimization, recent work has shifted toward feedforward reconstruction, where large networks regress Gaussian parameters or related scene representations from a few input images. Existing feedforward 3DGS or Gaussian reconstruction models target image pairs, sparse multi-view inputs, single-view or object-centric generation, free-view synthesis, pose-free inputs, and large reconstruction models~\cite{charatan2023pixelsplat,chen2024mvsplat,chen2024mvsplat360,wang2024freesplat,zhang2025transplat,zou2024triplane,tang2024lgm,zhang2024gs,hong2025pfplat,hong2024pf3plat}. Recent systems further improve depth coupling, unconstrained-view generalization, universal world reconstruction, and driving-specific surround-view reconstruction~\cite{xu2025depthsplat,jiang2025anysplat,liu2025worldmirror,tian2025drivingforward}. While feedforward methods are appealing for their single-pass global scene estimate, they lack a native per-scene optimization loop. Consequently, once a pretrained predictor outputs an imperfect scene, it cannot repair missing structures, gray uncertain regions, or over-smoothed textures. Adding more trajectory views does not necessarily solve this issue, as weak overlap and fixed network capacity complicate global fusion. We therefore utilize feedforward prediction as an initial reconstruction, focusing on how to repair it with local generative evidence without compromising global consistency.

\noindent\textbf{Diffusion Priors for Novel View Synthesis.}
Diffusion models provide strong 2D generative priors for hallucinating plausible details in unobserved regions~\cite{rombach2022high,brooks2023instructpix2pix}. This has inspired sparse-view novel view synthesis and 3D generation methods that synthesize additional viewpoints or view-consistent image sets~\cite{liu2023zero,liu2023syncdreamer,shi2024mvdream}, or distill 2D diffusion priors into 3D representations via score distillation, 3D-aware diffusion, iterative editing, and optimization~\cite{alldieck2024score,instructnerf2023,pmlr-v202-gu23a,karnewar2023holofusion}. Closest to our setting, recent methods enhance low-quality 3DGS renderings with view-consistent, video, single-step, or score-distillation diffusion priors, followed by 3D optimization or fine-tuning~\cite{liu20243dgs,wang2025fixinggs,wu2025difix3d+,wei2025gsfix3d}. While these approaches successfully recover high-frequency details and fill missing content, they are primarily designed for optimization-based representations. To maintain stability, existing methods often restrict diffusion-based repair to views very close to the observations, severely limiting their repair range in sparse driving trajectories. Additionally, generic 2D diffusion does not guarantee 3D consistency across independently generated frames. Disagreeing pseudo targets can make 3D optimization average incompatible evidence or create spurious structures, causing blur, ghosting, and repeated artifacts. A very recent feedforward method, DGGT~\cite{chen2025dggt}, also incorporates diffusion into its architecture, but uses it mainly for 2D post-processing of rendered images instead of repairing the 3D Gaussians.

\noindent\textbf{Controllable Novel View Video Synthesis and Reprojection Cues.}
Recent controllable novel view video synthesis methods show that camera trajectories can be redirected by combining generative video priors with geometric reprojection cues~\cite{Yu_2025_ICCV}. Related view-consistent generation work also studies how to identify low-certainty or inconsistent generated content under multi-view constraints~\cite{tao2024lcgen}. These results suggest that unprojection and reprojection can indicate which generated content is supported by source observations. Unlike methods that directly produce a 2D video, such as~\cite{chen2025dggt}, we use reprojection to evaluate diffusion-enhanced local pseudo targets before they supervise an explicit 3D Gaussian scene. This converts local generative outputs into confidence-weighted 3D training signals, retaining diffusion flexibility while avoiding unconditional reliance on hallucinated regions.

\noindent\textbf{Robust Optimization and Confidence Estimation.}
Multi-view consistency has long been used in classical multi-view geometry and MVS to detect unreliable observations, where reprojection and photometric errors help handle occlusions, outliers, and non-Lambertian effects~\cite{schonberger2016pixelwise}. Similar ideas appear in robust losses, robust NeRFs, and reconstruction pipelines that downweight inconsistent pixels or views during optimization~\cite{barron2019general,li2024rustnerf}. Confidence-weighted learning is also widely used in self-training and pseudo-labeling, including heteroscedastic confidence formulations~\cite{kendall2017uncertainties}. Recent 3DGS methods further introduce confidence-aware or depth, normal, or feature consistency regularization to stabilize geometry under sparse observations~\cite{chung2024depth,turkulainen2025dn,li2024dngaussian,huang2025fatesgs,zhang2025cdgs,huang2025det}. Existing methods usually reweight photometric objectives or mask training signals, but rarely intervene in the structural update rules of explicit 3D representations under generative pseudo supervision. We adapt these principles to diffusion-assisted feedforward repair. Instead of treating generated views as ground truth, we evaluate diffusion-enhanced pseudo-targets via deterministic reprojection-based cross-validation against support images, which yields a dense confidence map. By learning propagating this confidence to reweight the photometric loss and modulate both gradients and densification during 3DGS refinement, \myMethod suppresses unreliable generative residuals from spawning spurious Gaussians while empowering local details to improve the feedforward results.

%% file: sections/03_method.tex
\section{Methodology}
\label{sec:method}

\subsection{Feedforward Repair Problem}
\label{sec:preliminaries}

\noindent\textbf{Background.}
3DGS~\cite{kerbl3Dgaussians} represents a scene with a set of anisotropic Gaussians (mean $\mu$, covariance $\Sigma$, color $c$, opacity $\alpha$), renders pixel color via depth-sorted $\alpha$-blending, and updates topology through densification and pruning driven by accumulated 2D position gradients (we follow standard formulations and detail them in Appendix~\ref{sec:gs_grad}). Pre-trained image-to-image diffusion models $\mathcal{F}_{\text{diff}}$ further provide generative priors that map a low-quality rendering $\hat{I}$ to a high-fidelity target $I^*=\mathcal{F}_{\text{diff}}(\hat{I},c_{\text{cond}})$, where $c_{\text{cond}}$ denotes optional conditioning.

\noindent\textbf{Problem Setting.}
Given a sparse set of support views $\mathcal{S}$ with ground-truth RGB images $\{I_s^{\text{GT}}\}_{s\in\mathcal{S}}$ and cameras $\{(K_v, T_v)\}_{v=1}^{V}$, our goal is to optimize a 3DGS scene $\Theta$ that faithfully reconstructs the support views while rendering high-fidelity, geometrically consistent novel views.

\noindent\textbf{Why Feedforward Repair Is Different.} A feedforward 3DGS backbone produces an initial scene in a single forward pass. Since this prediction is a deterministic output of a pre-trained network, simply re-running the model cannot fix its errors. Repair must instead operate downstream in the Gaussian parameter space, making novel view supervision the main available lever. A natural strategy is to generate diffusion-enhanced pseudo-targets $I_n^*$ for novel views $n \notin \mathcal{S}$ using $\mathcal{F}_{\text{diff}}$, and optimize $\Theta$ by matching renderings $\hat{C}_n$ to them. However, independently generated diffusion targets can contain multi-view inconsistent hallucinations. Directly minimizing $\mathcal{L}(\hat{C}_n, I_n^*)$ may therefore corrupt the 3D representation, causing warped geometry and floating artifacts under conflicting 2D supervision. ConFixGS thus treats diffusion outputs as candidate evidence rather than ground truth.

\subsection{ConFixGS Overview}
\label{sec:overview}

\begin{figure}[t!]
   \centering
   \includegraphics[width=0.98\textwidth]{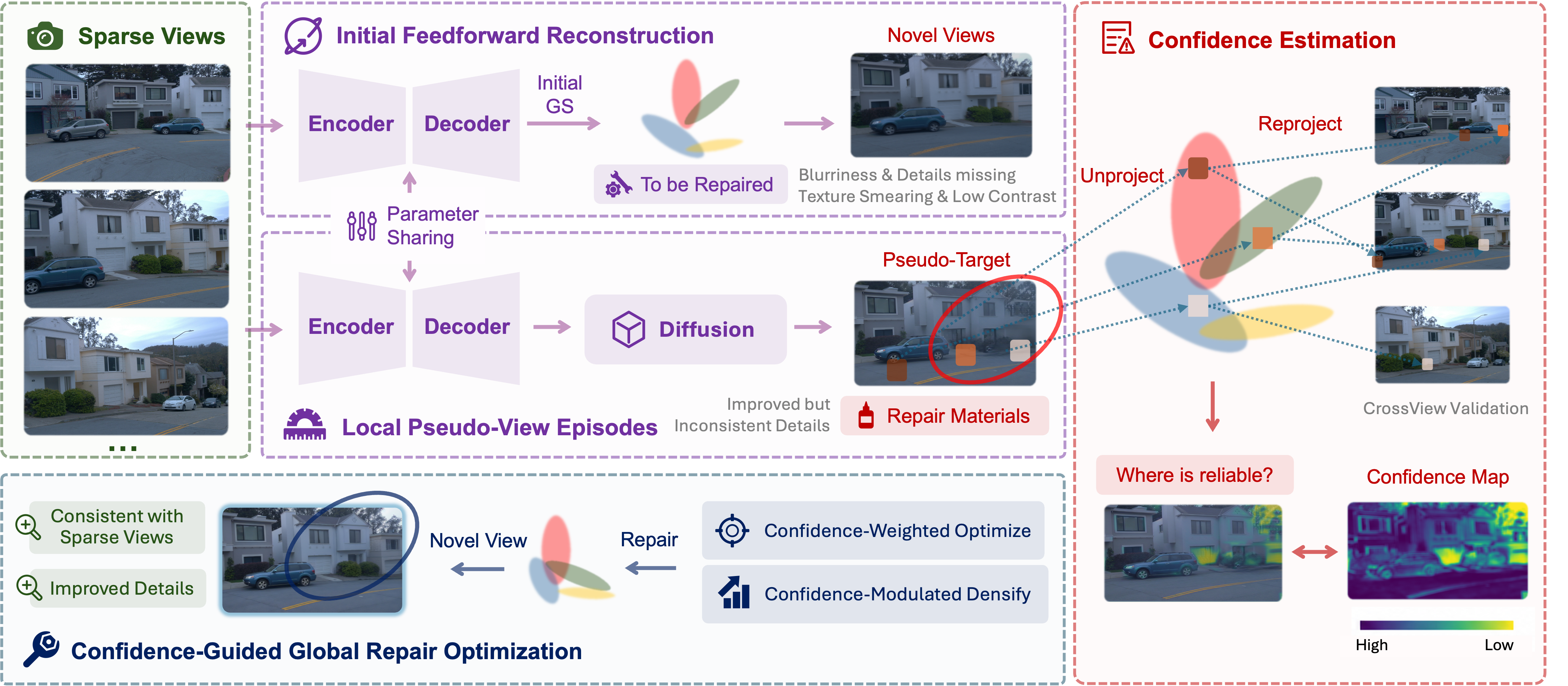}
   \caption{\textbf{ConFixGS Framework Overview.} ConFixGS consists of three stages: (\emph{i}) initial feedforward reconstruction with local pseudo-view episodes, (\emph{ii}) input-observation-guided confidence estimation through reprojection, and (\emph{iii}) confidence-modulated global repair optimization.}
    \label{fig:system}
   \vspace{-0.4cm}
\end{figure}

ConFixGS repairs a feedforward reconstruction by asking which locally generated details can be supported by the input observations (Fig.~\ref{fig:system}). The pipeline consists of three stages:
(\emph{i}) take the original feedforward 3DGS prediction $\mathcal{G}_0$ as the initial reconstruction to be repaired and construct detail-rich local pseudo-view episodes using local feedforward rendering and diffusion;
(\emph{ii}) compute a dense confidence map $W_n$ for each pseudo view by reprojecting generated pixels into the ground-truth support images, using $\mathcal{G}_0$ purely as a geometric scaffold for establishing correspondences;
(\emph{iii}) refine the global 3D Gaussians with a confidence-weighted objective, where low-confidence pixels are suppressed in both photometric gradients and standard 3DGS densification signals.

\subsection{Initial Feedforward Reconstruction and Local Pseudo-View Episodes}
\label{sec:stage1}

\noindent\textbf{Initial Feedforward Reconstruction.}
We employ a pre-trained feedforward 3DGS backbone $F_{\text{global}}$ to produce the initial reconstruction from sparse support views:
\begin{equation}
\label{eq:global_init}
\mathcal{G}_0 = F_{\text{global}}(\{I_s^{\text{GT}}\}_{s\in\mathcal{S}}).
\end{equation}
The predicted scene is a set of $N_0$ Gaussians
$\mathcal{G}_0 = \{(\mu_i, s_i, q_i, o_i, c_i)\}_{i=1}^{N_0}$,
where $\mu_i \in \mathbb{R}^3$ is the 3D mean, $s_i \in \mathbb{R}^3_{+}$ is the per-axis scale, $q_i \in \mathbb{R}^4$ is the rotation quaternion, $o_i \in [0,1]$ is the opacity, and $c_i \in \mathbb{R}^3$ is the RGB color. During Stage~\ref{sec:stage3} we adopt the standard logit/log reparameterization $\hat{o}_i = \sigma^{-1}(o_i)$ and $\hat{s}_i = \log s_i$ so that activated values stay in their valid ranges throughout training. This is the feedforward result we aim to repair: it can be over-smoothed or incomplete under sparse driving views, but it provides a useful initial state for later global refinement. Rendering $\mathcal{G}_0$ at any viewpoint $v$ yields a base RGB image $\hat{I}_v^{(0)}$, an accumulated opacity map $\hat{\alpha}_v^{(0)}$, and a depth proxy $d_v^{(0)}$. The base RGB is consumed by the local pseudo-view branch below, while the opacity and depth are reused as a geometric scaffold for confidence estimation in Sec.~\ref{sec:stage2}.

\noindent\textbf{Local Pseudo-View Episodes.}
For each novel view $n \notin \mathcal{S}$, we build a local episode around nearby trajectory observations and generate a pseudo-target $I_n^*$ by enhancing a local feedforward rendering conditioned on temporal context $I_n^{\text{ctx}}$ and the base render $\hat{I}_n^{(0)}$. This local formulation preserves the empirical advantage of small, highly overlapping inputs: local predictions may lack full-scene coverage, but their nearby novel views are often cleaner and easier to repair than a single global feedforward output. Following~\cite{wu2025difix3d+, chen2025dggt}, we use SD-Turbo as the generative prior:
\begin{equation}
\label{eq:pseudo_target}
I_n^* = \mathcal{F}_{\text{SD-Turbo}}\!\left(F_{\text{local}}(I_n^{\text{ctx}}, \hat{I}_n^{(0)})\right), \quad \forall n \notin \mathcal{S}.
\end{equation}
For support views, we preserve absolute fidelity by setting $I_s^* = I_s^{\text{GT}}$, yielding a complete target set $\{I_v^*\}_{v=1}^{V}$.

\subsection{Confidence Estimation via Input-Observation Reprojection}
\label{sec:stage2}

Novel view pseudo-targets may include diffusion hallucinations. We compute a reprojection-based confidence score from the intuition that a reliable pseudo-target pixel should be photo-consistent with the ground truth support views to the global feedforward model. The decision of whether a pseudo-target pixel is credible is grounded in the support-view RGB evidence; the initial reconstruction $\mathcal{G}_0$ enters this estimation only as a geometric scaffold, where the rendered depth $d_n^{(0)}$ provides a cheap proxy for establishing correspondences, and its accumulated opacity $\hat{\alpha}_n^{(0)}$ is used as a coverage indicator that flags regions where this scaffold is too sparse to support any reliable correspondence.

\noindent\textbf{Cross-View Unprojection.}
For a pixel $p=(u_p,v_p)$ in a novel view $n$, we unproject it into 3D using the depth proxy $d_n^{(0)}(p)$:
\begin{equation}
\label{eq:unprojection}
X(p) = R_n \left( d_n^{(0)}(p)\, K_n^{-1}\tilde{p} \right) + t_n,
\end{equation}
where $\tilde{p}=[u_p,v_p,1]^\top$ and $(R_n,t_n)$ is the camera-to-world transformation. We reproject $X(p)$ to a support view $s\in\mathcal{S}$:
\begin{equation}
\label{eq:reprojection}
p_s = \mathrm{dehom}\left( K_s R_s^{\top}(X(p) - t_s) \right).
\end{equation}

\noindent\textbf{Consensus Color and Discrepancy.}
Let $\mathcal{V}(p)\subset\mathcal{S}$ be the set of support views where the reprojection is valid (positive depth, within image bounds). We sample the support images and compute a reprojection consensus color:
\begin{equation}
\label{eq:reproj_consensus}
\bar{I}_{\text{reproj}}(p) = \frac{1}{|\mathcal{V}(p)| + \epsilon}\sum_{s\in \mathcal{V}(p)} \mathrm{Sample}(I_s^{\text{GT}}, p_s).
\end{equation}
We define the discrepancy between the pseudo-target and the consensus as
\begin{equation}
\label{eq:reproj_error}
e_n(p) = \frac{1}{3}\sum_{c\in\{R,G,B\}} \left| I_n^*(p,c) - \bar{I}_{\text{reproj}}(p,c)\right|.
\end{equation}

\noindent\textbf{Support-Validated Confidence.}
We define a support-validated pixel-wise confidence score $\tilde{w}_n(p)$:
\begin{equation}
\label{eq:raw_confidence}
\tilde{w}_n(p) =
\begin{cases}
\exp\!\left(-\frac{e_n(p)^2}{2\sigma_e^2}\right), & \text{if } \mathcal{V}(p)\neq\emptyset \text{ and } \hat{\alpha}_n^{(0)}(p)\ge\alpha_{\min}, \\
\gamma, & \text{if } \mathcal{V}(p)=\emptyset \text{ and } \hat{\alpha}_n^{(0)}(p)\ge\alpha_{\min}, \\
0, & \text{if } \hat{\alpha}_n^{(0)}(p)<\alpha_{\min},
\end{cases}
\end{equation}
where $\sigma_e$ controls the confidence-decay bandwidth, $\gamma$ is a conservative baseline confidence score assigned to regions whose correspondences are valid in geometry but cannot be photometrically validated by any support image, and the third branch zeros out pixels whose underlying scaffold $\mathcal{G}_0$ has no opacity coverage: there the depth proxy $d_n^{(0)}$ is unreliable, so the reprojection itself, and any photometric comparison built on it, is meaningless. We then apply spatial smoothing to obtain the final \emph{per-pixel confidence score}:
\begin{equation}
\label{eq:confidence_smoothing}
w_n(p) = (G_k * \tilde{w}_n)(p)\ \in [0,1], \qquad W_n \in [0,1]^{H\times W}, \quad (W_n)_p = w_n(p),
\end{equation}
where $G_k$ is a $k\times k$ mean filter. We refer to the resulting continuous-valued field $W_n$ as the \emph{confidence map} (in contrast to a binary segmentation mask), and to the scalar $w_n(p)$ as the per-pixel confidence score. For support views, we set the confidence map to all-ones:
\begin{equation}
\label{eq:support_confidence}
W_s = \mathbf{1}, \qquad \forall s \in \mathcal{S}.
\end{equation}

\subsection{Confidence-Guided Global Repair Optimization}
\label{sec:stage3}

With support-validated confidence maps $\{W_v\}_{v=1}^{V}$ and fixed targets $\{I_v^*\}_{v=1}^{V}$, we initialize $\Theta \leftarrow \mathcal{G}_0$ and refine 3DGS by optimizing a confidence-weighted objective. This stage is a lightweight repair step rather than a full reconstruction from scratch: the original feedforward prediction supplies the initial state, and the optimization selectively assimilates only reliable local pseudo-view evidence.

\noindent\textbf{Confidence-Weighted Objective.}
For a rendered image $\hat{C}_v$ and its target $I_v^*$, we define a confidence-weighted $L_1$ loss:
\begin{equation}
\label{eq:weighted_l1}
\mathcal{L}^{L_1}_{\text{tgt}}(v) = \frac{\sum_{p} w_v(p)\, \|\hat{C}_v(p) - I_v^*(p)\|_1}{\sum_{p} w_v(p) + \epsilon}.
\end{equation}
To apply pixel-wise weighting within SSIM while preserving local normalization, we use a blending approximation:
\begin{equation}
\label{eq:weighted_ssim}
\tilde{C}_v = w_v \odot \hat{C}_v + (1 - w_v) \odot I_v^*, \qquad
\mathcal{L}^{\text{SSIM}}_{\text{tgt}}(v) = 1 - \mathrm{SSIM}(\tilde{C}_v, I_v^*).
\end{equation}
The total target loss is
\begin{equation}
\label{eq:target_loss}
\mathcal{L}_{\text{tgt}}(v) = (1-\lambda_s)\mathcal{L}^{L_1}_{\text{tgt}}(v) + \lambda_s \mathcal{L}^{\text{SSIM}}_{\text{tgt}}(v).
\end{equation}
For support views, we additionally apply an anchoring loss $\mathcal{L}_{\text{gt}}(s)$ with weight $\lambda_{gt}$ to prevent drift.

\noindent\textbf{Confidence-Modulated Densification.}
We extend the standard 3DGS topology update by using the per-pixel confidence score to gate densification. Specifically, the normalized confidence weight $\lambda_v(p)$ modulates both the photometric gradient and the accumulated 2D position-gradient signal used for clone/split decisions. Low-confidence pixels are therefore prevented from driving either parameter updates or densification, while standard clone/split/prune thresholds remain unchanged. Details are provided in Appendix~\ref{sec:gs_grad}, with full pseudocode in Algorithm~\ref{alg:pipeline}.

%% file: sections/05_exp.tex
\section{Experiments}

\subsection{Setup}
\label{sec:setup}

\noindent\textbf{Dataset.} We evaluate our framework on the Waymo Open Dataset~\cite{mei2022waymo}, nuScenes~\cite{caesar2020nuscenes}, and KITTI~\cite{Voigtlaender2019CVPR}. To evaluate under sparse-view conditions, we use every 5th view as input, and the task is to synthesize the 4 intermediate views. Each scene contains 50 frames, unless the maximum sequence length is shorter.

\noindent\textbf{Metrics.} To assess photometric reconstruction quality, we report Peak Signal-to-Noise Ratio (PSNR), Structural Similarity Index Measure (SSIM)~\cite{wang2004image}, and Learned Perceptual Image Patch Similarity (LPIPS)~\cite{dosovitskiy2016generating}. For perceptual and temporal consistency, we report Fréchet Inception Distance (FID)~\cite{heusel2017gans} and Fréchet Video Distance (FVD)~\cite{unterthiner2018towards}.

\noindent\textbf{Baselines.} We benchmark against state-of-the-art feedforward 3DGS methods (pixelSplat~\cite{charatan2023pixelsplat}, MVSplat~\cite{chen2024mvsplat}, TranSplat~\cite{zhang2025transplat}) and driving-specific models (DrivingGaussian~\cite{zhou2024drivinggaussian}).
To demonstrate our method's versatility as a plug-and-play enhancement module, we integrate it atop two recent feedforward architectures: DepthSplat~\cite{xu2025depthsplat} and WorldMirror~\cite{liu2025worldmirror}. Following~\cite{wu2025difix3d+}, we also apply a diffusion model to further refine the novel view renderings, denoting this enhanced variant as ConFixGS+.
Each scene is repaired with only $K{=}1{,}000$ Adam iterations on top of the initial feedforward prediction; the full set of hyperparameters, learning rates, and densification schedule is reported in Appendix~\ref{sec:impl}.

\subsection{Quantitative Results}
\label{sec:quantitative}

\noindent\textbf{Photometric Reconstruction Quality.}
As shown in Tab.~\ref{tab:benchmark}, \myMethod consistently improves existing feedforward architectures. ConFixGS-DS and ConFixGS-WM substantially outperform their baselines, DepthSplat and WorldMirror, with PSNR gains of up to \textbf{3.30 dB} and \textbf{3.68 dB} on Waymo, respectively. \myMethod also improves cross-dataset generalization: on KITTI, ConFixGS-DS raises DepthSplat from below AnySplat to above it, achieving 14.48 dB better than 14.08 dB PSNR.

\noindent\textbf{Perceptual and Temporal Consistency.}
Tab.~\ref{tab:fidfvd} evaluates perceptual quality with FID and FVD. Although diffusion-based post-processing slightly degrades standard photometric metrics in Tab.~\ref{tab:benchmark}, it substantially improves perceptual realism and temporal consistency. This reflects the perception-distortion trade-off, where pixel-level metrics penalize plausible high-frequency details introduced by diffusion. For example, ConFixGS-DS+ nearly halves the FID on KITTI compared with ConFixGS-DS (151.18 vs. 285.02) and achieves the best or second-best FVD scores across all benchmarks.

\noindent\textbf{Comparison with Difix3D+~\cite{wu2025difix3d+}.}
Tab.~\ref{tab:difix_comp} compares \myMethod with Difix3D+~\cite{wu2025difix3d+} on Waymo. \myMethod consistently outperforms the original Difix3D+ baseline across key metrics. To isolate the source of improvement, we further compare against a modified Difix3D+ using the same WorldMirror feedforward initialization as our pipeline. Even under this controlled setting, \myMethod maintains a clear quantitative advantage, showing that the gains come from the proposed support-validated confidence module rather than a stronger geometric initialization.

\begin{table*}[t]
\centering
\caption{\textbf{Quantitative evaluation on the Waymo, nuScenes, and KITTI datasets.} The best, second-best, and third-best results are highlighted in \textbf{bold}, \underline{underline}, and \textit{italics}, respectively. This will be used for the following tables.}
\resizebox{\textwidth}{!}{
\begin{tabular}{l ccc ccc ccc}
\toprule
\multirow{2}{*}{Method} & \multicolumn{3}{c}{Waymo} & \multicolumn{3}{c}{nuScenes} & \multicolumn{3}{c}{KITTI} \\
\cmidrule(lr){2-4} \cmidrule(lr){5-7} \cmidrule(lr){8-10}
& PSNR $\uparrow$ & SSIM $\uparrow$ & LPIPS $\downarrow$
& PSNR $\uparrow$ & SSIM $\uparrow$ & LPIPS $\downarrow$
& PSNR $\uparrow$ & SSIM $\uparrow$ & LPIPS $\downarrow$ \\
\midrule
pixelSplat~\cite{charatan2023pixelsplat}    & 20.20 & 0.6043 & 0.2818 & 16.23 & 0.3018 & 0.5267 & 12.70 & 0.2456 & \underline{0.5004} \\
MVSplat~\cite{chen2024mvsplat}            & 13.64 & 0.3849 & 0.6478 & 12.66 & 0.2072 & 0.6643 & 7.14  & 0.0467 & 0.6980 \\
TranSplat~\cite{zhang2025transplat}       & 15.34 & 0.4605 & 0.5125 & 15.17 & 0.3144 & 0.6126 & 8.08  & 0.1138 & 0.6922 \\
DepthSplat~\cite{xu2025depthsplat}        & \textit{24.56} & \textit{0.7990} & \textit{0.2550} & \textit{16.96} & \textit{0.5057} & \textit{0.5224} & 12.52 & \textit{0.3190} & 0.5863 \\
DrivingForward~\cite{tian2025drivingforward} & 20.63 & 0.6765 & 0.3174 & 14.24 & 0.3422 & 0.5354 & 8.71  & 0.2644 & 0.5914 \\
WorldMirror~\cite{liu2025worldmirror}     & 20.38 & 0.6000 & 0.4949 & 15.07 & 0.3621 & 0.6136 & 9.13  & 0.1855 & 0.6652 \\
AnySplat~\cite{jiang2025anysplat}         & 20.49 & 0.6683 & 0.3159 & 15.85 & 0.4486 & 0.5384 & \textit{14.08} & 0.3117 & \textbf{0.4719} \\
\midrule
ConFixGS-WM                                  & 24.06 & 0.7193 & 0.4176 & 16.60 & 0.3448 & 0.5679 & 12.27 & 0.2056 & 0.6323 \\
ConFixGS-WM+                                 & 22.88 & 0.6808 & 0.3631 & 16.10 & 0.3125 & 0.5400 & 12.31 & 0.1997 & 0.5891 \\
ConFixGS-DS                                  & \textbf{27.86} & \textbf{0.8670} & \textbf{0.2060} & \textbf{19.37} & \textbf{0.5469} & \underline{0.4585} & \textbf{14.48} & \textbf{0.3456} & 0.5475 \\
ConFixGS-DS+                                 & \underline{25.39} & \underline{0.8138} & \underline{0.2540} & \underline{19.05} & \underline{0.5209} & \textbf{0.4451} & \underline{14.20} & \underline{0.3259} & \textit{0.5332} \\
\bottomrule
\end{tabular}
}
\label{tab:benchmark}
\end{table*}

\begin{table*}[t]
\begin{minipage}{0.48\linewidth}
  \centering
  \caption{\textbf{Quantitative evaluation of perceptual and temporal consistency} across the Waymo, nuScenes, and KITTI datasets.}
  \label{tab:fidfvd}
  \resizebox{\linewidth}{!}{
    \begin{tabular}{l cc cc cc}
    \toprule
    \multirow{2}{*}{Method} & \multicolumn{2}{c}{Waymo} & \multicolumn{2}{c}{nuScenes} & \multicolumn{2}{c}{KITTI} \\
    \cmidrule(lr){2-3} \cmidrule(lr){4-5} \cmidrule(lr){6-7}
    & FID $\downarrow$ & FVD $\downarrow$
    & FID $\downarrow$ & FVD $\downarrow$
    & FID $\downarrow$ & FVD $\downarrow$ \\
    \midrule
    WorldMirror~\cite{liu2025worldmirror}     & 185.3 & 90.3  & 324.0  & 160.2  & 327.1  & 249.5 \\
    DepthSplat~\cite{xu2025depthsplat}        & 68.60 & \textit{24.38} & 297.99 & 178.02 & 308.74 & 231.64 \\
    \midrule
    ConFixGS-WM                                  & 92.3  & 36.4  & 191.6  & \textit{67.5}   & 298.6  & 232.9 \\
    ConFixGS-WM+                                 & \textit{57.2}  & 25.9  & \textbf{116.8}  & \textbf{50.9}   & \underline{168.3}  & \underline{128.4} \\
    ConFixGS-DS                                  & \underline{50.02} & \underline{17.26} & \textit{168.37} & 70.04  & \textit{285.02} & \textit{197.17} \\
    ConFixGS-DS+                                 & \textbf{48.46} & \textbf{14.96} & \underline{119.15} & \underline{53.00} & \textbf{151.18} & \textbf{110.26} \\

    \bottomrule
    \end{tabular}
  }
\end{minipage}\hfill
\begin{minipage}{0.49\linewidth}
  \centering
  \caption{\textbf{Comparison against Difix3D+}~\cite{wu2025difix3d+} on the Waymo dataset.}
  \label{tab:difix_comp}
  \resizebox{\linewidth}{!}{
    \begin{tabular}{l ccccc}
    \toprule
    Method & PSNR$\uparrow$ & SSIM$\uparrow$ & LPIPS$\downarrow$ & FID$\downarrow$ & FVD$\downarrow$ \\
    \midrule
    Difix3D        & \textit{21.12} & \textbf{0.7817} & 0.5065 & 269.64 & 169.23 \\
    Difix3D+       & 19.04 & \textit{0.6856} & 0.4914 & 167.85 & 139.08 \\
    Difix3D$^*$    & 20.25 & 0.6094 & 0.4790 & \textit{85.89} & 47.86 \\
    Difix3D$^*+$   & 19.46 & 0.5713 & \textit{0.4375} & \underline{79.80} & \textit{42.67} \\
    ConFixGS         & \underline{24.06} & \underline{0.7193} & \underline{0.4176} & 92.26 & \underline{36.39} \\
    ConFixGS+        & \textbf{22.88} & 0.6808 & \textbf{0.3631} & \textbf{57.20} & \textbf{25.88} \\
    \bottomrule
    \end{tabular}
  }
  \vspace{0.15cm}
  \par\raggedright\scriptsize
  * denotes the Difix3D baseline is a modified feedforward version.
\end{minipage}
\end{table*}

\subsection{Ablation Study}
\label{sec:ablation}

To validate our design, we evaluate three degraded variants on nuScenes in Tab.~\ref{tab:ablation_nuscenes_novel}. \emph{w/o Diffusion} extracts $I^*$ directly from the local or $\mathcal{G}_0$ rendering, removing diffusion-based enhancement. \emph{w/o Local Confidence} disables the support-validated confidence map and uses diffusion-enhanced pseudo-targets with uniform unit weight, allowing unfiltered pseudo supervision during global optimization. \emph{w/o Global Opt} removes the final 3DGS optimization and relies on independent local outputs, which are less likely to be globally consistent.

Tab.~\ref{tab:ablation_nuscenes_novel} confirms the contribution of each component. \emph{w/o Diffusion} retains competitive PSNR, e.g., 18.91 in ConFixGS+, but substantially degrades perceptual quality, with FID increasing to 235.0, reflecting the perception-distortion trade-off. \emph{w/o Local Confidence} reduces PSNR from 19.03 to 17.46 in ConFixGS, showing that uniformly weighting diffusion-enhanced pseudo-targets introduces 3D conflicts and harms geometric fidelity. \emph{w/o Global Opt} leads to the worst consistency, with FVD increasing to 159.3, demonstrating the need for global 3D optimization to maintain coherence.

\begin{figure}[t!]
   \centering
   \includegraphics[width=\textwidth]{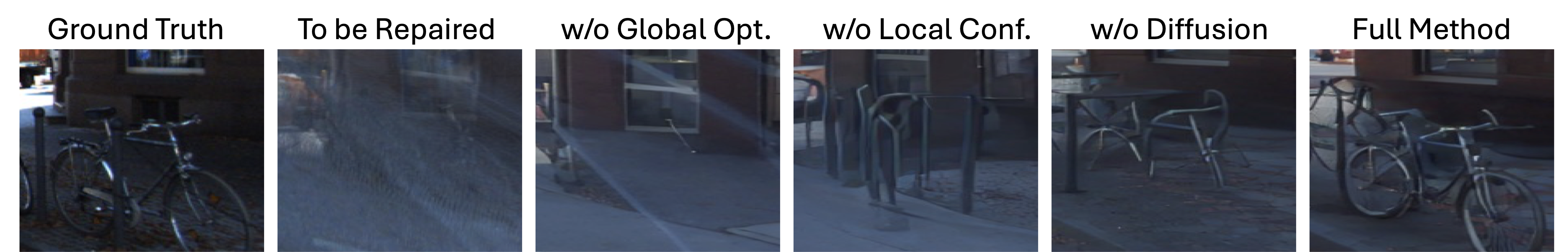}
   \caption{\textbf{Visual ablation study.} We visualize the effects of individual components by removing one module at a time. We use WorldMirror~\cite{liu2025worldmirror} as the feedforward backbone.}
    \label{fig:vis_ablation}
\end{figure}

Fig.~\ref{fig:vis_ablation} illustrates the role of each module. Without diffusion, under-observed regions miss high-frequency details, such as the incomplete bicycle. Without local confidence gating, unreliable pseudo-targets introduce hallucinated structures, such as cabinet-like walls and billboard-like window patterns unsupported by input views. Without global optimization, the result stays close to the feedforward baseline, WorldMirror in our implementation, since local diffusion improvements are not integrated into a consistent 3D scene. Together, diffusion completes details, confidence gating suppresses hallucinations, and global optimization consolidates reliable pseudo-targets.

\begin{table}[t]
\centering
\caption{\textbf{Ablation study on nuScenes.} We compare metrics in both ConFixGS and ConFixGS+ modes.}
\label{tab:ablation_nuscenes_novel}
\resizebox{\linewidth}{!}{%
\begin{tabular}{l ccccc ccccc}
\toprule
\multirow{2}{*}{Method} & \multicolumn{5}{c}{ConFixGS} & \multicolumn{5}{c}{ConFixGS+} \\
\cmidrule(lr){2-6} \cmidrule(lr){7-11}
 & PSNR$\uparrow$ & SSIM$\uparrow$ & LPIPS$\downarrow$ & FID$\downarrow$ & FVD$\downarrow$
 & PSNR$\uparrow$ & SSIM$\uparrow$ & LPIPS$\downarrow$ & FID$\downarrow$ & FVD$\downarrow$ \\
\midrule
Full Method (Ours)
 & \textbf{19.03} & \underline{0.5252} & \textbf{0.4602} & \underline{166.9} & \textbf{65.4}
 & \underline{18.84} & \underline{0.4934} & \underline{0.4468} & \textbf{112.3} & \textbf{45.4} \\
w/o Diffusion
 & \underline{18.86} & \textbf{0.5407} & \underline{0.4846} & \textit{235.0} & \underline{87.6}
 & \textbf{18.91} & \textbf{0.5087} & \textbf{0.4399} & \underline{121.7} & \underline{47.6} \\
w/o Confidence Guidance
 & \textit{17.46} & 0.4587 & \textit{0.4921} & \textbf{161.6} & \textit{95.5}
 & \textit{17.22} & 0.4195 & 0.5013 & \textit{134.3} & \textit{76.4} \\
w/o Global Opt
 & 17.01 & \textit{0.4792} & 0.5281 & 316.3 & 159.3
 & 17.15 & \textit{0.4394} & \textit{0.4947} & 147.5 & 90.6 \\
\bottomrule
\end{tabular}%
}
\end{table}

\subsection{Qualitative Analysis}
\label{sec:qualitative}
\begin{figure}[t!]
   \centering
   \includegraphics[width=\textwidth]{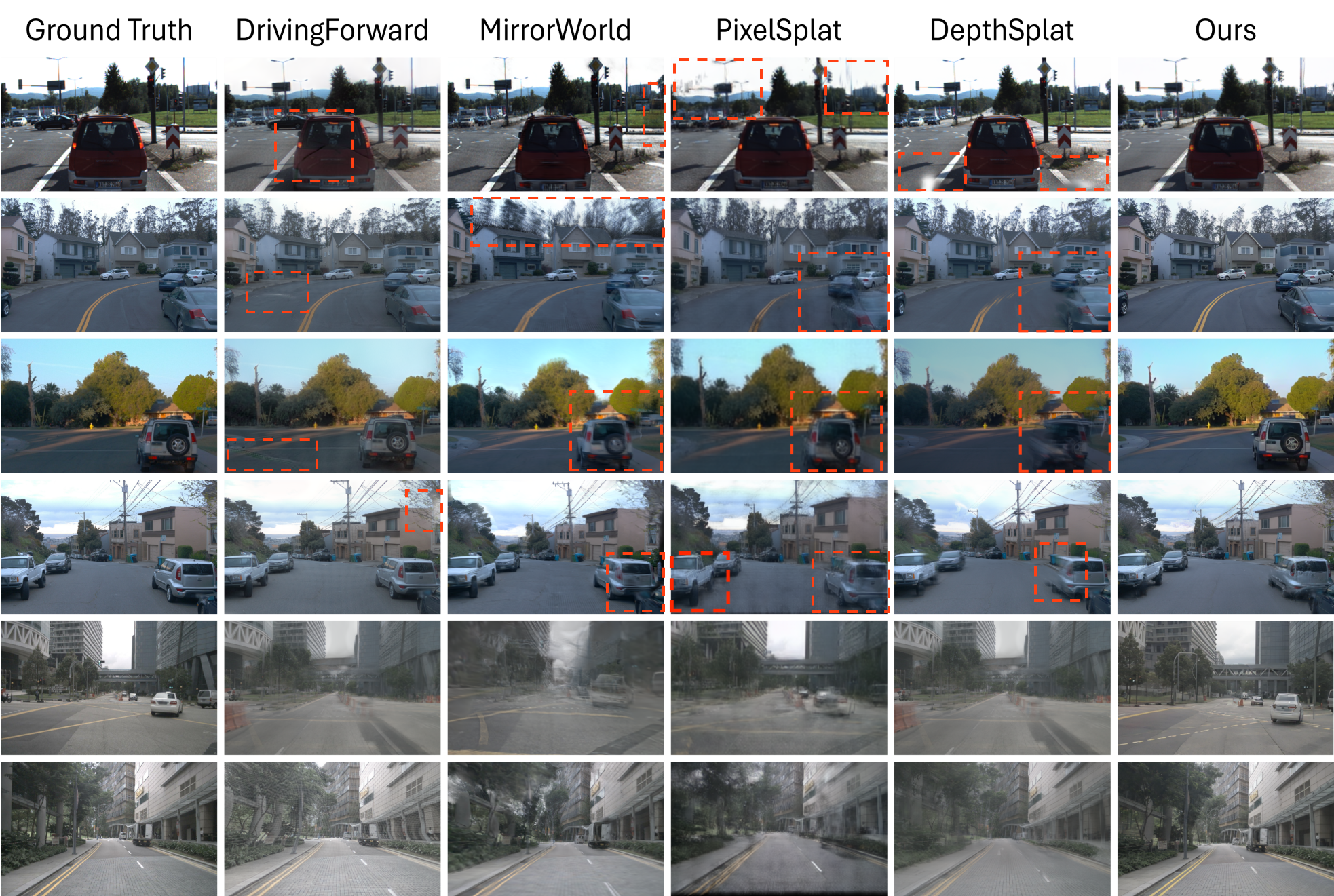}
   \caption{Visual comparison. We use DepthSplat~\cite{xu2025depthsplat} as the feedforward backbone for our approach. More results are shown in Figs.~\ref{fig:vis_comp_depthsplat-1} and \ref{fig:vis_comp_depthsplat-2}, with additional backbone comparisons in \ref{fig:fix_comp_hunyuan}, \ref{fix_comp_anysplat}, and
        \ref{fig:fix_comp_drivfwd}.}
    \label{fig:vis_comp}
\end{figure}

Fig.~\ref{fig:teaser} left shows \myMethod with WorldMirror as the feedforward backbone. Compared with the original results, our method reduces blur, ghosting, and unstable textures, yielding clearer structures and more coherent appearance. The diffusion prior enhances high-frequency details in under-observed regions, while the local feedforward branch supplies complementary geometry and object cues through fine-grained local Gaussians. This helps recover missing or inaccurate content, such as an incomplete bicyclist or a waiting vehicle at a traffic light. More results are provided in Sec.~\ref{sec:add_comp}.

Fig.~\ref{fig:vis_comp} compares \myMethod with state-of-the-art baselines across multiple datasets and driving scenes. \myMethod produces higher-quality novel view renderings with fewer artifacts, sharper details, better completeness, and stronger fidelity to the ground truth. It remains stable across diverse scene layouts and observation sparsity, avoiding common baseline failure modes.

%% file: sections/06_conclusion.tex
\section{Conclusion}

In summary, we present ConFixGS, a confidence-guided repair approach specifically designed to fix feedforward 3DGS in sparse-view driving scenes. ConFixGS uses a training-free reprojection check to cross-validate diffusion-enhanced pseudo-targets against ground-truth support images, producing dense per-pixel confidence maps without auxiliary predictors. These maps guide lightweight per-scene optimization by modulating photometric supervision and Gaussian densification, enabling the model to absorb reliable generative details while suppressing hallucinated or multi-view inconsistent evidence. More broadly, our results suggest that geometry-grounded confidence estimation offers a simple yet effective principle for routing generative priors into 3D Gaussian representations, without relying on iterative diffusion-optimization loops.

\noindent\textbf{Limitations.}
ConFixGS is a confidence-guided repair framework for feedforward 3DGS, and therefore works best when reliable evidence exists in either the global prediction or the local feedforward reconstruction. In severely under-constrained regions, stronger teacher signals, such as denser anchors or temporally consistent priors, may further enhance recovery. Our reprojection-based confidence also assumes cross-view consistency, leaving view-dependent effects and fast-moving objects as interesting directions for future extension.

%% file: sections/07_imp_details.tex
\section{Implementation Details}
\label{sec:impl}

\noindent\textbf{Datasets and protocol.}
We evaluate on driving scenes from Waymo Open Dataset~\cite{mei2022waymo}, nuScenes~\cite{caesar2020nuscenes}, and KITTI~\cite{Voigtlaender2019CVPR}. For each scene, we extract 50 consecutive frames from the front-facing camera, using every 5th frame as a support (context) view (10 views) and the remaining 40 frames as novel views, yielding an extremely sparse-view setting (80\% novel views). In comparison, prior driving novel view synthesis benchmarks often use protocols with much denser context views, e.g., holding out every 4th frame~\cite{zhou2024drivinggaussian,sun2025splatflow,yang2024emernerf,song2025coda} (25\% novel views) or every 2nd frame~\cite{tian2025drivingforward} (50\% novel views) as test views. All images are resized with aspect ratio preserved (longer side 784 px), with the height rounded to multiples of 14 for compatibility with the 14×14 patch size of the DINOv2 ViT-L/14 in WorldMirror~\cite{liu2025worldmirror}. For the DepthSplat~\cite{xu2025depthsplat} backbone, images are also resized to match the pretrained model and the encoder’s 14×14 patch size. We use dataset-provided camera intrinsics and poses, and additionally apply COLMAP-based scale calibration~\cite{schoenberger2016sfm,schoenberger2016mvs}.

\noindent\textbf{Global-local feedforward backbones.}
Our framework is plug-and-play w.r.t.\ the feedforward initializer $\mathcal{G}_0$. We report results using two representative global architectures: DepthSplat~\cite{xu2025depthsplat} and WorldMirror~\cite{liu2025worldmirror}. For practical compatibility, DepthSplat is executed via a lightweight subprocess bridge, while WorldMirror runs in-process at the same target resolution. The local feedforward operator $F_{\text{local}}$ in Eq.~\eqref{eq:pseudo_target} is \emph{not} a separate network: it shares the same backbone weights as the global model $F_{\text{global}}$, and is obtained by re-running the same feedforward predictor over a small local window of nearby trajectory views centered on each novel view $n$. In Sec.~\ref{sec:further_backbone}, we further provide results using DrivingForward~\cite{tian2025drivingforward} and AnySplat~\cite{jiang2025anysplat} as alternative backbones, along with both qualitative and quantitative comparisons before and after enhancement, to demonstrate the generalizability of our approach across diverse feedforward initializers.

\noindent\textbf{Diffusion refinement module.}
For image-space generative enhancement, we adopt SD-Turbo~\cite{sauer2024adversarial} as our diffusion prior, applied to refine novel view renderings from local feedforward models, and produce pseudo-targets $I^*$. We rescale the diffusion outputs back to the rendering resolution and retain the camera parameters in corresponding Gaussians.

\noindent\textbf{Support-validated reprojection confidence.}
A core component is our support-validated confidence estimation for novel views. Using the depth proxy $d_n^{(0)}$ rendered from the initial feedforward reconstruction $\mathcal{G}_0$ as a geometric scaffold, we unproject diffusion-refined pixels into 3D and reproject them into multiple support views to measure photometric disagreement with the input observations. This disagreement is converted into the per-pixel confidence score with $\sigma_e=0.10$; pixels that cannot be validated by any support image receive the conservative baseline confidence score $\gamma=0.3$, and pixels falling in regions whose accumulated opacity $\hat{\alpha}_n^{(0)}$ is below the scaffold-coverage threshold $\alpha_{\min}=0.3$ are zeroed out, since their depth proxy and the corresponding reprojection are unreliable. We then apply spatial smoothing with a $15\times 15$ mean-filter kernel ($k=15$) to obtain the final confidence map $W_n$.

\noindent\textbf{Confidence-weighted global optimization.}
We optimize the global 3DGS for 1k iterations using Adam, with learning rates $1.6{\times}10^{-4}$ for Gaussian positions and $5{\times}10^{-3}$ for spherical harmonics; we set $\lambda_s=0.2$ and $\lambda_{gt}=1.0$. During training, the confidence map gates distillation gradients on novel views and also modulates densification to prevent low-confidence regions from spawning spurious Gaussians. Densification is performed every 100 steps under confidence control.

\noindent\textbf{Hardware and software environment.} All experiments are conducted on a single NVIDIA L40S GPU (46 GiB VRAM) hosted on a workstation with two AMD EPYC 9354 32-core CPUs and ~750 GB system RAM, running Ubuntu 22.04 (Linux kernel 5.15). ConFixGS is implemented in PyTorch 2.4 with CUDA 12.4. To preserve dependency parity with each released baseline, every comparison method is executed inside its own original conda environment; cross-environment communication (e.g., the DepthSplat feedforward call) is handled through a lightweight subprocess bridge.

\begin{algorithm}[H]
\caption{Support-Validated Confidence Optimization}
\label{alg:pipeline}
\begin{algorithmic}[1]
\State \textbf{Input:} Support GT images $\{I_s^{\text{GT}}\}_{s\in\mathcal{S}}$, cameras $\{(K_v,T_v)\}_{v=1}^{V}$, hyperparameters $\sigma_e,\gamma,\alpha_{\min},\lambda_s,\lambda_{gt},T,T_d$
\State \textbf{Output:} Optimized 3D scene $\mathcal{G}_{\text{final}}$
\vspace{0.05in}

\State $\mathcal{G}_0 \gets F_{\text{global}}(\{I_s^{\text{GT}}\}_{s\in\mathcal{S}})$ \Comment{Initial feedforward reconstruction, Eq.~\eqref{eq:global_init}}
\State $(\hat{I}_v^{(0)}, \hat{\alpha}_v^{(0)}, d_v^{(0)}) \gets \mathrm{Render}(\mathcal{G}_0; K_v, T_v), \ \forall v$ \Comment{Base render, opacity, depth (Sec.~\ref{sec:stage1})}

\For{$n \notin \mathcal{S}$}
    \State $I_n^* \gets \mathrm{LocalEpisode}(I_n^{\text{ctx}}, \hat{I}_n^{(0)}, \mathcal{F}_{\text{SD-Turbo}})$
    \Comment{Pseudo-target generation, Eq.~\eqref{eq:pseudo_target}}
\EndFor
\State $I_s^* \gets I_s^{\text{GT}}, \ \forall s \in \mathcal{S}$ \Comment{Support targets are GT}

\vspace{0.05in}
\For{$n \notin \mathcal{S}$}
    \State $W_n \gets \mathrm{ReprojConfidence}\!\left(I_n^*, d_n^{(0)}, \hat{\alpha}_n^{(0)}, \{I_s^{\text{GT}}\}_{s\in\mathcal{S}}\right)$
    \Comment{Support-validated confidence map via Sec.~\ref{sec:stage2}, Eq.~\eqref{eq:raw_confidence}}
\EndFor
\State $W_s \gets \mathbf{1}, \ \forall s \in \mathcal{S}$ \Comment{All-ones confidence map for support views}

\vspace{0.05in}
\State Initialize parameters $\Theta \gets \mathcal{G}_0$
\For{iteration $t=1,\dots,T$}
    \State Sample mini-batch $\mathcal{B} \subset \{1,\dots,V\}$, let $\mathcal{B}_{gt} = \mathcal{B} \cap \mathcal{S}$
    \State $\hat{C}_v \gets \mathrm{Render}(\Theta; K_v, T_v), \ \forall v \in \mathcal{B}$
    \Comment{Current renderings}

    \State $\mathcal{L} \gets \frac{1}{|\mathcal{B}|}\sum_{v\in\mathcal{B}} \mathcal{L}_{\text{tgt}}(v)$
    \Comment{Weighted objective, Eqs.~\eqref{eq:weighted_l1}--\eqref{eq:target_loss}}

    \If{$|\mathcal{B}_{gt}| > 0$}
        \State $\mathcal{L} \gets \mathcal{L} + \lambda_{gt}\frac{1}{|\mathcal{B}_{gt}|}\sum_{s\in\mathcal{B}_{gt}}\mathcal{L}_{\text{gt}}(s)$
        \Comment{GT anchoring (Sec.~\ref{sec:stage3})}
    \EndIf

    \State Backpropagate and update $\Theta$ via Adam
    \Comment{Gradients are modulated by $W_v$ through Appendix Eq.~\eqref{eq:app_weighted_grad}}

    \If{$t \bmod T_d == 0$}
        \State Densify/Prune Gaussians using the standard 3DGS strategy
        \Comment{Topology updates follow standard 3DGS thresholds; densification signals are weighted by Appendix Eq.~\eqref{eq:app_weighted_densification}}
    \EndIf
\EndFor
\State \Return $\mathcal{G}_{\text{final}} \gets \Theta$
\end{algorithmic}
\end{algorithm}

%% file: sections/08_opt_details.tex
\section{Confidence-Weighted Gradients and Objective}
\label{sec:gs_grad}

To mitigate both geometric and photometric degradation caused by diffusion hallucinations, we propose a confidence-weighted optimization scheme guided by support-validated, discrepancy-based confidence maps.
For each novel view pixel, we reproject the pseudo-target into the ground-truth support images used by the global feedforward model and compute its photometric discrepancy from the reprojected multi-view consensus; pixels with high disagreement receive a low confidence score.
This section details the formulations governing our modulated backward pass. We demonstrate how this per-pixel confidence score acts as an explicit gate that embeds input-observation consistency into both the continuous parameter optimization and the discrete densification of the repaired 3DGS scene.

\subsection{Explicit Coupling in the Backward Pass}
{Rather than unconditionally backpropagating the 2D photometric residual, our framework modulates the gradient using the support-validated confidence map $W_v$, whose per-pixel confidence score $w_v(p) = (G_k * \tilde{w}_v)(p)$ is obtained by spatially smoothing the raw confidence score $\tilde{w}_v$ with a mean filter $G_k$. By expanding the normalized weight $\lambda_v(p) = \frac{w_v(p)}{\sum_{p'} w_v(p')}$, the confidence-weighted gradient with respect to any 3D Gaussian parameter $\theta_i \in \{\mu_i, \hat{s}_i, q_i, \hat{o}_i, c_i\}$ is explicitly formulated as:}
\begin{equation}
\frac{\partial \mathcal{L}}{\partial \theta_i} = \frac{1}{|\mathcal{B}|} \sum_{v \in \mathcal{B}} \sum_{p} { \underbrace{ \left[ \frac{(G_k * \tilde{w}_v)(p)}{\sum_{p'} w_v(p')} \right] }_{\text{Confidence Gating }\lambda_v(p)} } \cdot \underbrace{\nabla_{\hat{C}}\ell_v(p)}_{\text{2D Loss Gradient}} \cdot \underbrace{\frac{\partial \hat{C}_v(p)}{\partial \theta_i}}_{\text{Rendering Derivative}}.
\label{eq:app_weighted_grad}
\end{equation}
where the raw per-pixel confidence score $\tilde{w}_v(p)$ encodes support-image validation through the valid support view set $\mathcal{V}(p)$, the photometric discrepancy $e_v(p) = \frac{1}{3}\sum_{c\in\{R,G,B\}} | I_v^*(p,c) - \bar{I}_{\text{reproj}}(p,c) |$ between the pseudo-target and the reprojected multi-view consensus, and the geometric coverage indicator $\hat{\alpha}_v^{(0)}(p)$ rendered from the initial feedforward reconstruction $\mathcal{G}_0$:
\begin{equation}
{
\tilde{w}_v(p) =
\begin{cases}
\exp\!\left(-\frac{e_v(p)^2}{2\sigma_e^2}\right), & \text{if } \mathcal{V}(p) \neq \emptyset \text{ and } \hat{\alpha}_v^{(0)}(p) \ge \alpha_{\min}, \\
\gamma, & \text{if } \mathcal{V}(p) = \emptyset \text{ and } \hat{\alpha}_v^{(0)}(p) \ge \alpha_{\min}, \\
0, & \text{if } \hat{\alpha}_v^{(0)}(p) < \alpha_{\min}.
\end{cases}
}
\end{equation}
{This expanded formulation reveals the essence of our method: the parameter update is modulated by the target's consistency with the photometric consensus established by the support views, while the initial reconstruction $\mathcal{G}_0$ enters only as a geometric scaffold for establishing correspondences. Note that in the main paper, we use $n$ to index novel views ($n \notin \mathcal{S}$); here we use $v$ as a generic view index over all $V$ views.}

\subsection{Mechanism of Densification Gating}
\label{app:densification_proof}
{Within our framework, the confidence-modulated 2D position gradient accumulation $\bar{g}_i$ for Gaussian $i$ across its associated views $S_i$ becomes deeply coupled with the support-validated per-pixel confidence score $w_v(p)$:}
\begin{equation}
\label{eq:app_weighted_densification}
{
\bar{g}_i = \frac{1}{|S_i|} \sum_{t \in S_i} \left\| \sum_{p} \lambda_{v_t}(p) \cdot \nabla_{\hat{C}}\ell_{v_t}(p) \cdot \frac{\partial \hat{C}_{v_t}(p)}{\partial \mu_i^{2D}} \right\|.
}
\end{equation}
{This explicit expansion illustrates three distinct gating behaviors dictated by the underlying raw confidence score $\tilde{w}_v(p)$:
\begin{enumerate}
    \item \textbf{Geometric Hallucinations ($\hat{\alpha}_v^{(0)}(p) < \alpha_{\min}$):} If the pixel falls in a region where the initial reconstruction $\mathcal{G}_0$ has no opacity coverage, the rendered depth $d_v^{(0)}(p)$ is unreliable, so any reprojection based on it---and the photometric comparison built on top---is meaningless. We set $\tilde{w}_v(p) \equiv 0$, mathematically zeroing out the positional gradient and preventing the creation of floaters in empty space.
    \item \textbf{Photometric Inconsistencies ($e_v(p) \gg 0$):} For regions with valid scaffold coverage but severe multi-view conflicts (where the pseudo-target diverges from the support-view consensus $\bar{I}_{\text{reproj}}$), the exponential decay forces $\tilde{w}_v(p) \to 0$, naturally suppressing conflicting gradient updates.
    \item \textbf{Unobserved Regions ($\mathcal{V}(p) = \emptyset$):} When the geometry is covered by $\mathcal{G}_0$ but no support view can validate pixel $p$ photometrically (e.g., occluded or out-of-view), we assign a baseline $\tilde{w}_v(p) = \gamma$ instead of zero. This preserves a moderate gradient flow, allowing the diffusion prior to guide expansion where multi-view consensus is unavailable, while remaining less influential than validated regions.
\end{enumerate}
Therefore, our support-validated confidence map suppresses gradient accumulation from unreliable regions with high reprojection discrepancy or with no scaffold coverage, reducing the likelihood that spurious artifacts trigger densification, while preserving limited gradient flow in unobserved-but-covered regions via $\gamma$. Note that $S_i$ denotes the set of views where Gaussian $i$ is visible, and $v_t$ is the view index of the $t$-th element in $S_i$, \ie, $v_t \in S_i$.}

%% file: sections/10_efficiency.tex
\section{Complexity and Efficiency Analysis}
\label{sec:efficiency}

We analyze the computational cost of \myMethod in a form aligned with the three-stage pipeline described in the Methodology section.
Given $K$ total views, including $K_s$ support views and $K_n = K-K_s$ novel views, the overall computation is dominated by three components:
(1) initial feedforward reconstruction and local diffusion,
(2) reprojection-based confidence estimation, and
(3) confidence-modulated global repair optimization.

\vspace{0.3em}
\noindent\textbf{Stage 1: Initial feedforward reconstruction and local diffusion.}
In the first stage, the feedforward backbone $F_{\text{global}}$ is executed once on the support views to predict the initial Gaussian scene $\mathcal{G}_0$ to be repaired, after which $\mathcal{G}_0$ is rendered at all $K$ camera poses to produce base renderings $\hat{I}_v^{(0)}$.
Then, for each novel view $n \notin \mathcal{S}$, the local branch $F_{\text{local}}$ and the diffusion prior $\mathcal{F}_{\text{SD-Turbo}}$ are used to generate the pseudo-target $I_n^*$.
Let $T_{\mathrm{ff}}^{g}$ and $T_{\mathrm{ff}}^{l}$ denote the costs of one global and one local feedforward call, respectively, $T_{\mathrm{render}}$ the cost of one 3DGS rendering, and $T_{\mathrm{diff}}$ the cost of one diffusion refinement.
The complexity of Stage~1 is therefore
\begin{equation}
T_{\mathrm{stage1}}
=
T_{\mathrm{ff}}^{g}
+
K\,T_{\mathrm{render}}
+
K_n\left(T_{\mathrm{ff}}^{l} + T_{\mathrm{diff}}\right).
\end{equation}
This stage scales linearly with the number of novel views, and in practice is typically dominated by the per-view rendering of $\mathcal{G}_0$, the repeated local feedforward, and the diffusion refinement. Empirically, on a representative Waymo scene (50 frames, single L40S GPU), Stage~1 takes $\sim$12.5\,min with the DepthSplat backbone (where the per-view DepthSplat call is dispatched as a subprocess bridge) and $\sim$6.4\,min with the WorldMirror backbone (where the model stays resident in process across calls).

\vspace{0.3em}
\noindent\textbf{Stage 2: Reprojection-based confidence estimation.}
For each novel view pixel, Stage~2 unprojects the pixel using the depth proxy $d_n^{(0)}$ rendered from the initial feedforward reconstruction $\mathcal{G}_0$ (a cheap geometric scaffold for correspondence), reprojects it into all support views, samples valid support observations, and computes the discrepancy-based confidence map $W_n$.
Assuming image resolution $H \times W$, the cost of reprojection-based confidence estimation is linear in the number of novel views, support views, and pixels:
\begin{equation}
T_{\mathrm{stage2}} = \mathcal{O}(K_n K_s H W).
\end{equation}
Since this stage involves geometric reprojection, image sampling, and photometric discrepancy computation, its cost is substantially smaller than feedforward inference or iterative 3DGS optimization in practice. In our setting ($K_n{=}40$, $K_s{=}10$, $448{\times}768$), Stage~2 finishes in $\sim$15\,s, accounting for less than $2\%$ of the total pipeline cost.

\vspace{0.3em}
\noindent\textbf{Stage 3: Confidence-modulated global repair optimization.}
In the final stage, we initialize $\Theta \leftarrow \mathcal{G}_0$ and optimize the Gaussian scene for $S$ iterations using mini-batches of size $B$.
Each iteration requires rendering the current scene for the sampled views, computing the confidence-weighted photometric objective, and back-propagating gradients to Gaussian parameters.
Let $N$ be the number of Gaussians.
The resulting complexity can be expressed as
\begin{equation}
T_{\mathrm{stage3}}
=
\mathcal{O}\!\left(
S \cdot B \cdot T_{\mathrm{render}} + S \cdot N
\right),
\end{equation}
where the rendering term captures forward rasterization and the $N$-dependent term reflects parameter updates and accumulated densification statistics.
Because confidence maps suppress gradients in unreliable regions, the effective optimization is concentrated on high-confidence pixels, which reduces unnecessary gradient competition and stabilizes densification. With $S{=}1{,}000$ and $B{=}4$ on $\sim$3.4\,M Gaussians, Stage~3 runs at $\sim$0.11\,s/step and completes in under $2$\,min, making it the second-largest contributor to total runtime after the local feedforward and base rendering of Stage~1.

\vspace{0.3em}
\noindent\textbf{Overall complexity.}
Combining the three stages, the total cost of \myMethod is
\begin{equation}
T_{\mathrm{total}}
=
T_{\mathrm{ff}}^{g}
+
K\,T_{\mathrm{render}}
+
K_n\left(T_{\mathrm{ff}}^{l} + T_{\mathrm{diff}}\right)
+
\mathcal{O}(K_n K_s H W)
+
\mathcal{O}\!\left(S \cdot B \cdot T_{\mathrm{render}} + S \cdot N \right).
\end{equation}
This expression highlights three important properties of our pipeline.
First, the method scales linearly with the number of novel views $K_n$, since pseudo-target construction and confidence estimation are both performed per novel view.
Second, the reprojection-based confidence estimation is lightweight compared to feedforward inference and global optimization.
Third, unlike iterative diffusion-refinement pipelines that repeatedly alternate between rendering and generative correction, our method performs pseudo-target generation only once before optimization, after which refinement proceeds entirely in 3D under confidence-aware supervision.
As a result, the overall efficiency is primarily governed by the chosen feedforward backbone and the number of global optimization steps.

\subsection{Empirical Wall-Clock Comparison}
\label{sec:efficiency_empirical}

To complement the theoretical analysis above, we report measured per-scene wall-clock and peak GPU memory in Table~\ref{tab:efficiency} on a representative Waymo scene (50 frames, $K_s{=}10$ support and $K_n{=}40$ novel views, $448{\times}768$ resolution) using a single NVIDIA L40S 46\,GiB GPU. Because confidence-modulated supervision concentrates gradients on confident regions and suppresses diffusion hallucinations early in optimization, \myMethod converges in $S{=}1{,}000$ steps. This is more than an order of magnitude fewer than the $S{=}30{,}000$ schedule used by per-scene 3DGS and Difix3D+~\cite{wu2025difix3d+} and is the dominant source of the observed speed-up.
As baselines we include a per-scene 3DGS reconstruction, and a Difix3D+~\cite{wu2025difix3d+} reproduction that interleaves $10$ rounds of diffusion fixing into the $30{,}000$-step 3DGS schedule. \emph{Test time} reports the wall-clock for synthesizing a single novel view using the final Gaussian scene; this cost is dominated by gsplat rasterization and PNG encoding to disk, and is therefore identical across methods.

\begin{table}[t]
\centering
\caption{\textbf{Per-scene wall-clock and peak GPU memory} on a representative Waymo scene (50 frames, $K_s{=}10$ support / $K_n{=}40$ novel, $448{\times}768$, single NVIDIA L40S 46\,GiB GPU). \emph{Train time} reports the full pipeline wall-clock; \emph{Test time} reports the wall-clock per synthesized novel view. Confidence-modulated supervision lets \myMethod converge in $S{=}1{,}000$ optimization steps with either the WorldMirror or DepthSplat backbone, more than $\boldsymbol{40{\times}}$ faster than the $S{=}30{,}000$ schedule of per-scene 3DGS or Difix3D+, while \myMethod-DS additionally uses noticeably less peak GPU memory than either baseline.}
\label{tab:efficiency}
\setlength{\tabcolsep}{4pt}
\renewcommand{\arraystretch}{1.15}
\small
\begin{tabular}{l c c c c c}
\toprule
\textbf{Method} & \textbf{Backbone} & \textbf{Opt.\ steps $S$} & \textbf{Train time}$\downarrow$ & \textbf{Test time}$\downarrow$ & \textbf{Peak GPU Mem.}$\downarrow$ \\
\midrule
Plain 3DGS                          & --                  & 30\,000                  & 9\,h\,39\,min  & 1.28\,s/view & 27\,GiB \\
Difix3D+~\cite{wu2025difix3d+}      & --                  & 30\,000 + 10 fix         & 10\,h\,39\,min & 1.28\,s/view & 29\,GiB \\
\midrule
\textbf{\myMethod-WM} (\emph{ours}) & WorldMirror         & 1\,000                   & \textbf{8\,min\,53\,s}  & 1.28\,s/view & 35\,GiB \\
\textbf{\myMethod-DS} (\emph{ours}) & DepthSplat          & 1\,000                   & \textbf{14\,min\,39\,s} & \textbf{1.28\,s/view} & \textbf{21\,GiB} \\
\bottomrule
\end{tabular}
\end{table}

\begin{table}[t]
\centering
\caption{\textbf{Stage-wise wall-clock breakdown of \myMethod-DS} (DepthSplat backbone, $S{=}1{,}000$ optimization steps, single L40S). Stage~1 dominates the total cost; Stage~2 (confidence estimation) is negligible; Stage~3 (confidence-modulated optimization) finishes in under two minutes because the optimizer starts from a near-converged $\mathcal{G}_0$.}
\label{tab:efficiency_stages}
\setlength{\tabcolsep}{4pt}
\renewcommand{\arraystretch}{1.15}
\small
\begin{tabular}{l l c c}
\toprule
\textbf{Stage} & \textbf{Component} & \textbf{Time} & \textbf{\% of pipeline} \\
\midrule
Stage 1a & $\mathcal{G}_0$ global feedforward                          & 53\,s        & 6.0\% \\
Stage 1b & Render $\mathcal{G}_0$ at all $K{=}50$ views                & 4\,m\,34\,s  & 31.2\% \\
Stage 1c & 50 local feedforward calls                                  & 6\,m\,55\,s  & 47.2\% \\
Stage 1d & SD-Turbo diffusion ($K_n{=}40$ novel views)                 & 9\,s         & 1.0\% \\
\multicolumn{2}{l}{\textbf{Stage 1 total} (initial reconstruction + pseudo-targets)} & \textbf{12\,m\,30\,s} & \textbf{85.4\%} \\
\midrule
Stage 2  & Reprojection-based confidence estimation                    & 15\,s        & 1.7\% \\
\midrule
Stage 3  & Confidence-modulated 3DGS optimization ($S{=}1{,}000$, $B{=}4$) & 1\,m\,53\,s & 12.9\% \\
\midrule
\multicolumn{2}{l}{\textbf{Pipeline total}}                            & \textbf{14\,m\,39\,s} & 100\% \\
\bottomrule
\end{tabular}
\end{table}

\noindent\textbf{Per-step supervision cost.}
\myMethod\ computes the pseudo-targets $I_n^*$ and the confidence maps $W_n$ once before optimization, in Stages~1d and~2, and applies them as cached tensors at every iteration of Stage~3; each \myMethod\ optimization step therefore performs only a rasterization, a per-pixel multiplication with the cached $W_n$, the photometric loss, and an Adam update. Difix3D+ instead refreshes its diffusion supervision through periodic \emph{fix events}, each of which re-renders all $K_n{=}40$ novel views and re-invokes the diffusion fixer subprocess end-to-end. Table~\ref{tab:efficiency_step} contrasts the wall-clock of one such step under each scheme on the same Waymo scene: a Difix3D+ fix-injection step is more than three orders of magnitude slower than a single \myMethod\ optimization step, which is why dense per-step diffusion supervision is feasible only when the supervision is precomputed and cached as in \myMethod.

\begin{table}[t]
\centering
\caption{\textbf{Per-step wall-clock} on the same Waymo scene (single L40S).}
\label{tab:efficiency_step}
\setlength{\tabcolsep}{8pt}
\renewcommand{\arraystretch}{1.2}
\small
\begin{tabular}{l c}
\toprule
\textbf{Step type} & \textbf{Per-step time}$\downarrow$ \\
\midrule
\myMethod\ optimization step                                    & $\boldsymbol{\sim\!115\,\text{ms}}$ \\
Difix3D+~\cite{wu2025difix3d+} step with diffusion-fix injection & $\sim\!180\,\text{s}$              \\
\bottomrule
\end{tabular}
\end{table}

\noindent\textbf{Why \myMethod is efficient.}
Three factors make \myMethod roughly $\boldsymbol{40{\times}}$ faster than per-scene 3DGS and $\boldsymbol{43{\times}}$ faster than Difix3D+ on the same scene, while consuming comparable or lower peak memory:
(\emph{i}) pseudo-target generation runs only \emph{once} before optimization, avoiding the expensive iterative render--diffuse--retrain loop of Difix3D+, where each fixing round adds another full pass through the 40 novel views;
(\emph{ii}) Stage~3 starts from a near-converged $\mathcal{G}_0$ and only needs to assimilate the trusted pseudo-view evidence, reducing the required optimization horizon by an order of magnitude (from $S{=}30{,}000$ to $S{=}1{,}000$);
(\emph{iii}) confidence-modulated densification suppresses gradient accumulation in unreliable regions, so the per-step rendering cost stays low throughout optimization rather than growing as densification produces spurious Gaussians under conflicting supervision.
The two backbones we evaluate exhibit complementary efficiency profiles: \myMethod-WM has shorter total wall-clock because the WorldMirror model stays resident in process across the $K_n$ local feedforward calls, but at higher peak GPU memory; \myMethod-DS dispatches DepthSplat as a per-call subprocess, which adds $\sim$7\,s of bridge overhead per call but keeps memory low and avoids carrying the model state across the whole pipeline. In both cases, Stage~2 (confidence estimation) and Stage~3 (confidence-modulated optimization) are negligible compared to Stage~1, confirming that the additional overhead of \myMethod beyond the chosen backbone is small and bounded.

%% file: sections/09_add_visual.tex
\section{Additional Comparisons and Results}
\label{sec:add_comp}

\begin{figure}[t!]
   \centering
   \includegraphics[width=\textwidth]{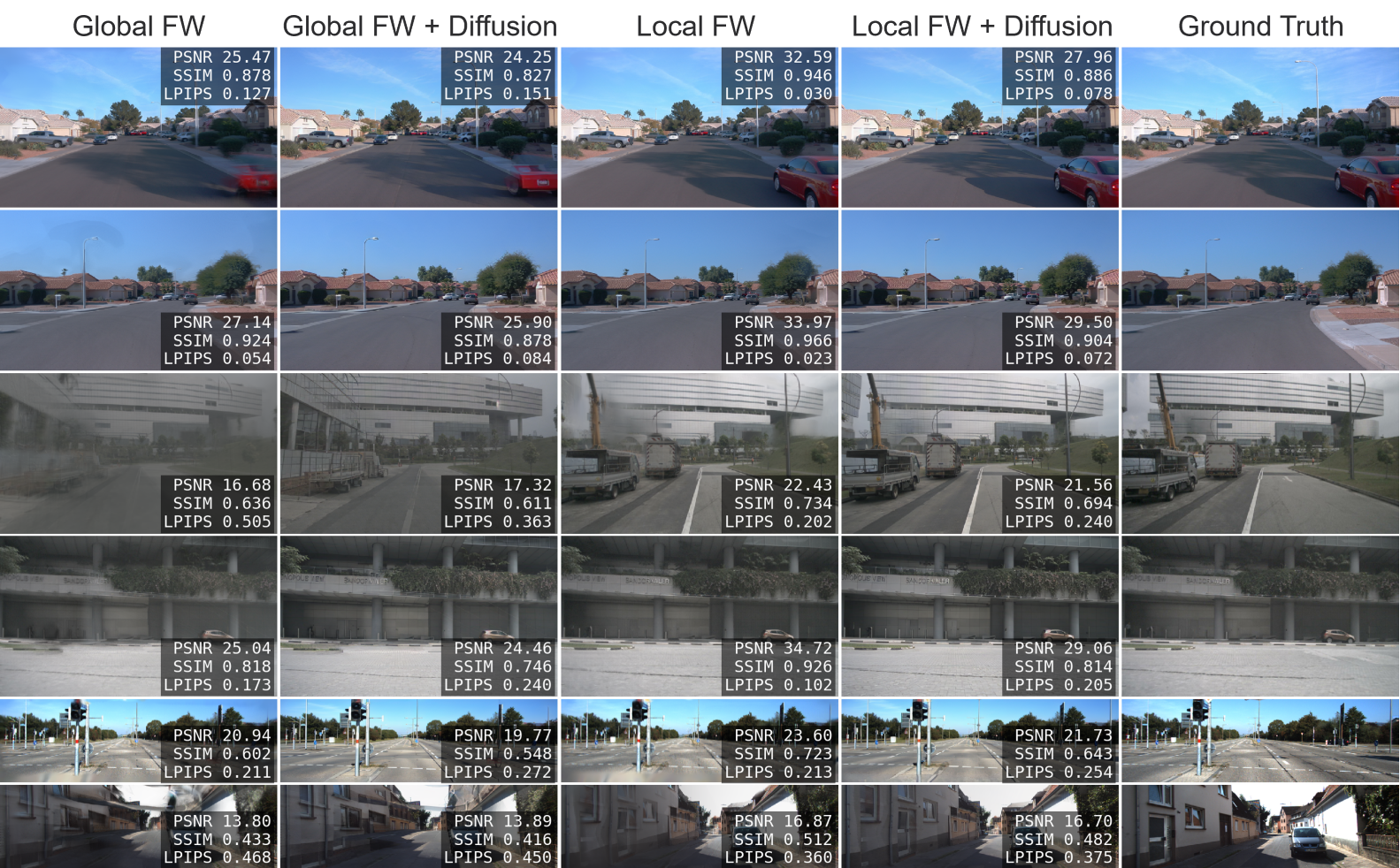}
   \caption{\textbf{Comparison of global and local feedforward 3DGS rendering on Waymo, nuScenes, and KITTI.} For each scene we compare the global feedforward (FW) rendering of $\mathcal{G}_0$, which has to reconstruct the entire trajectory from sparse, weakly overlapping support views, against the local FW rendering produced from a small subset around each novel view, and report the diffusion-enhanced version. Across all three datasets, local FW yields visibly sharper and more reliable novel views than global FW, and the diffusion-enhanced local renderings are the most repairable, supporting the key observation in Sec.~\ref{sec:intro} that motivates our local pseudo-view episode design.}
    \label{fig:local_vs_global_fw}
\end{figure}

\begin{figure}[t!]
   \centering
\includegraphics[width=\textwidth]{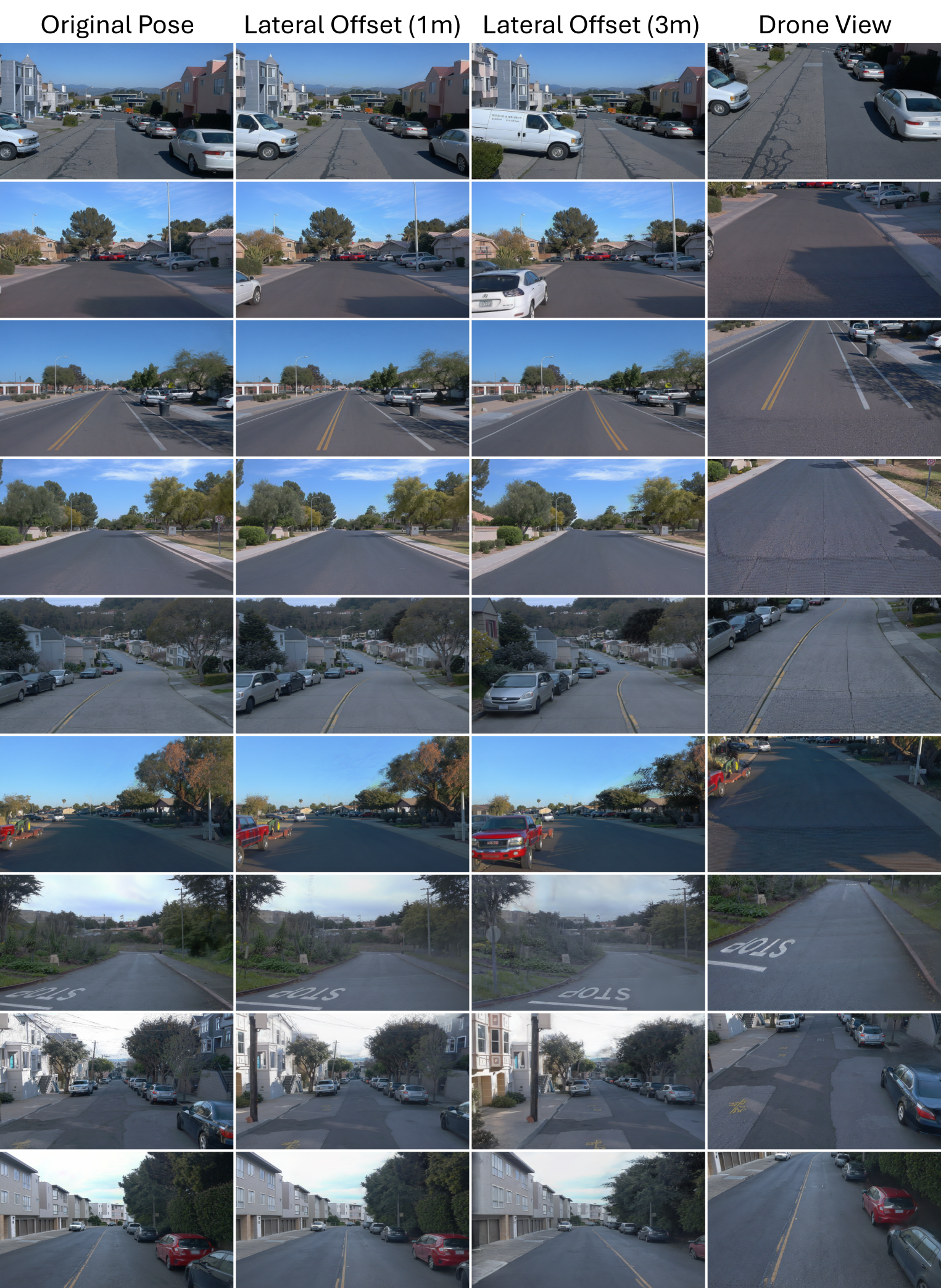}
   \caption{\textbf{Additional novel view synthesis results} under more challenging viewpoint shifts, including lateral offsets of 1\,m and 3\,m, as well as a drone-style viewpoint at approximately 2.5\,m height with a 20$^\circ$ downward pitch.}
    \label{fig:ood_traj}
   \vspace{-0.4cm}
\end{figure}

\begin{figure}[t!]
   \centering
\includegraphics[width=\textwidth]{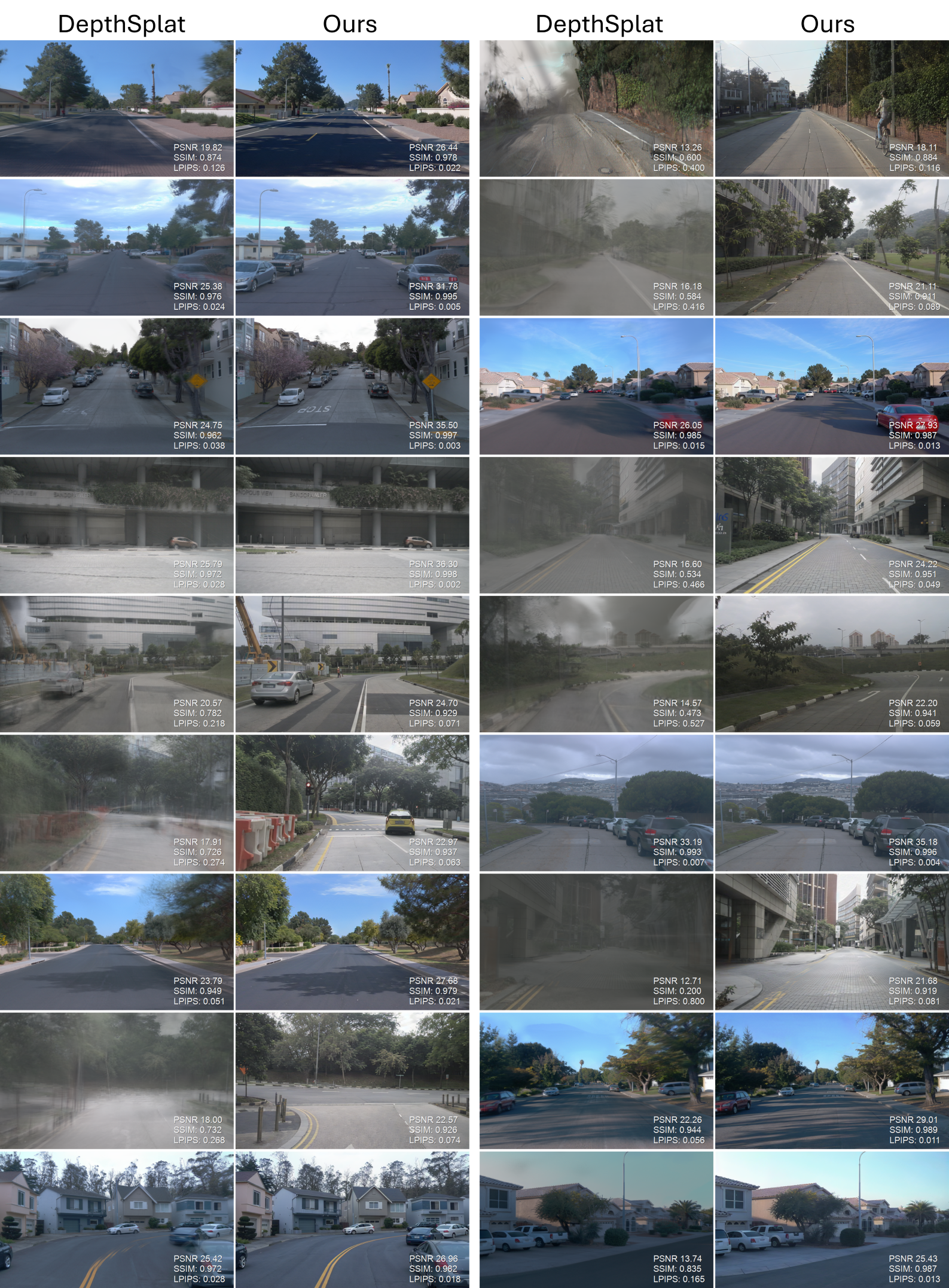}
   \caption{\textbf{Qualitative comparison of DepthSplat}~\cite{xu2025depthsplat} before and after \myMethod enhancement -- I.}
    \label{fig:vis_comp_depthsplat-1}
\end{figure}

\begin{figure}[t!]
   \centering
\includegraphics[width=\textwidth]{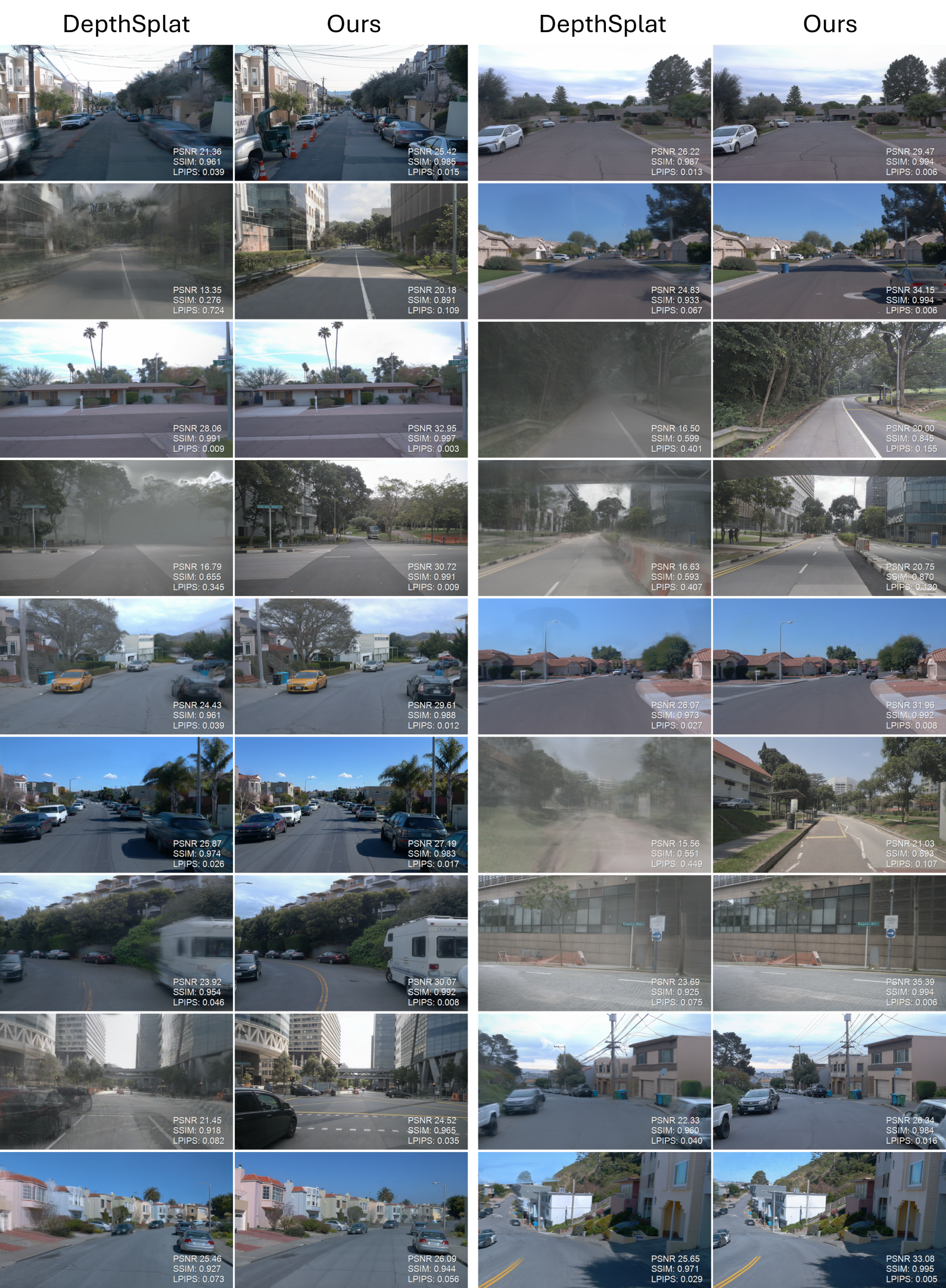}
   \caption{\textbf{Qualitative comparison of DepthSplat}~\cite{xu2025depthsplat} before and after \myMethod enhancement -- II.}
    \label{fig:vis_comp_depthsplat-2}
\end{figure}

\begin{figure}[t!]
   \centering
\includegraphics[width=\textwidth]{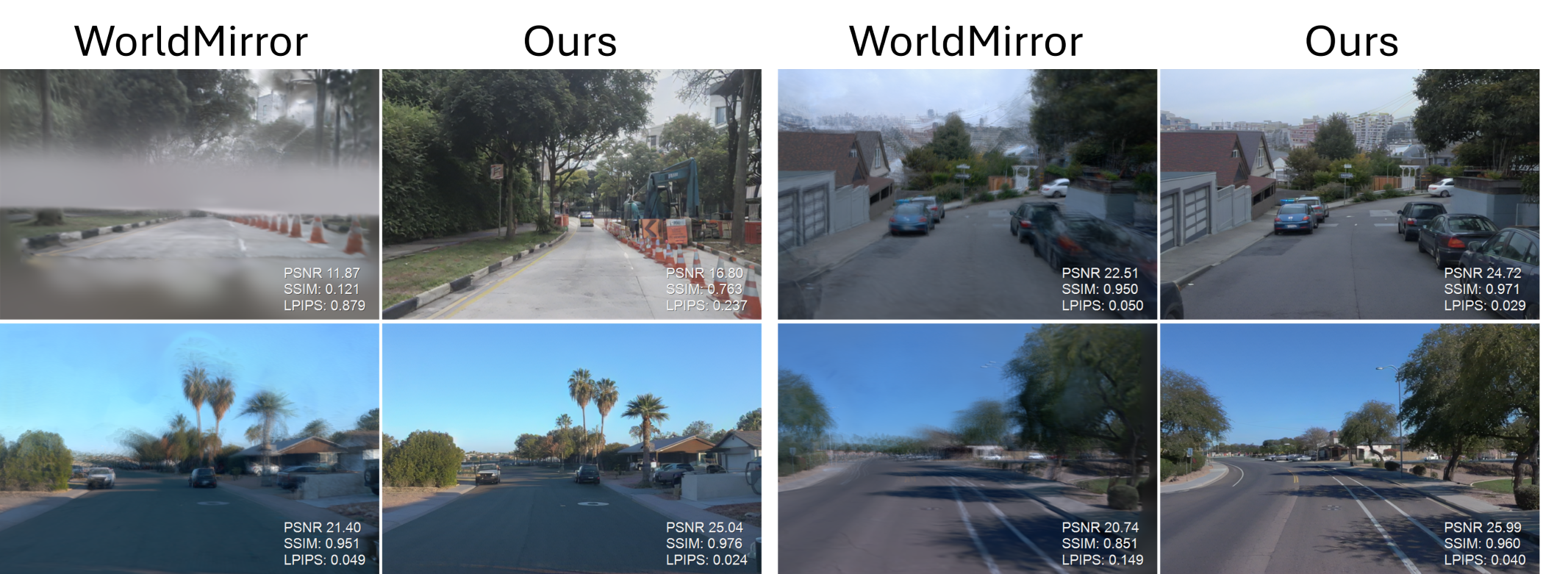}
\caption{\textbf{Qualitative comparison of WorldMirror}~\cite{liu2025worldmirror} before and after \myMethod enhancement.}
    \label{fig:fix_comp_hunyuan}
\end{figure}

\begin{figure}[t!]
   \centering
\includegraphics[width=\textwidth]{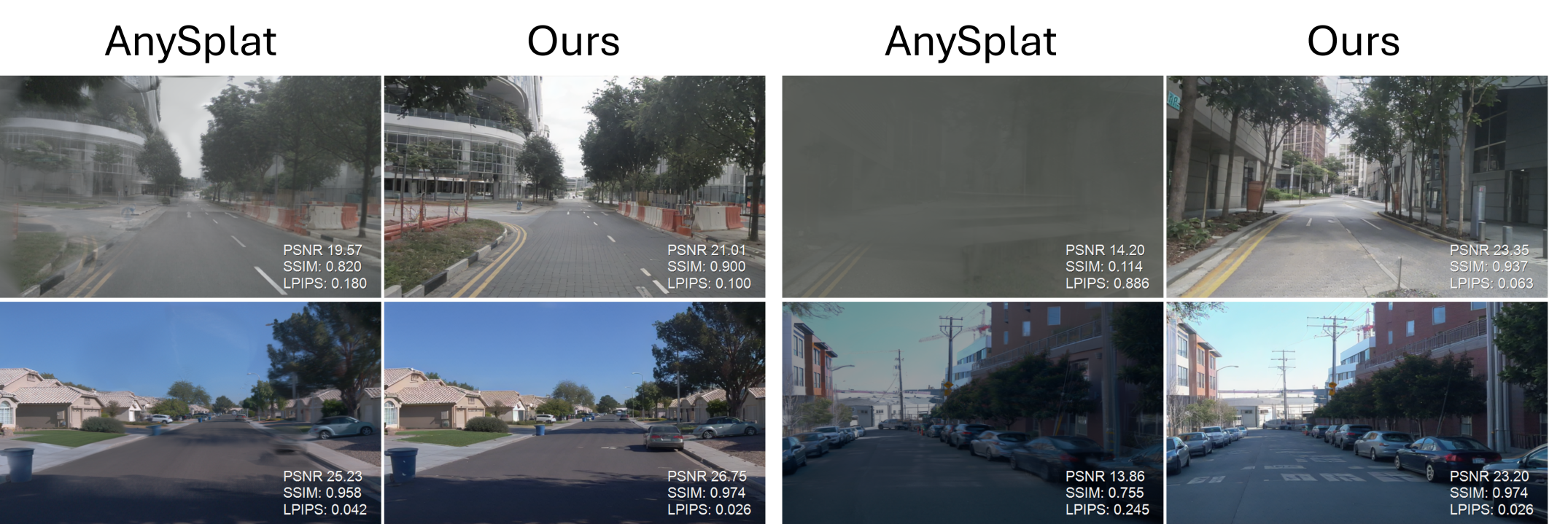}
\caption{\textbf{Qualitative comparison of AnySplat}~\cite{jiang2025anysplat} before and after \myMethod enhancement.}
    \label{fix_comp_anysplat}
\end{figure}

\begin{figure}[t!]
   \centering
\includegraphics[width=\textwidth]{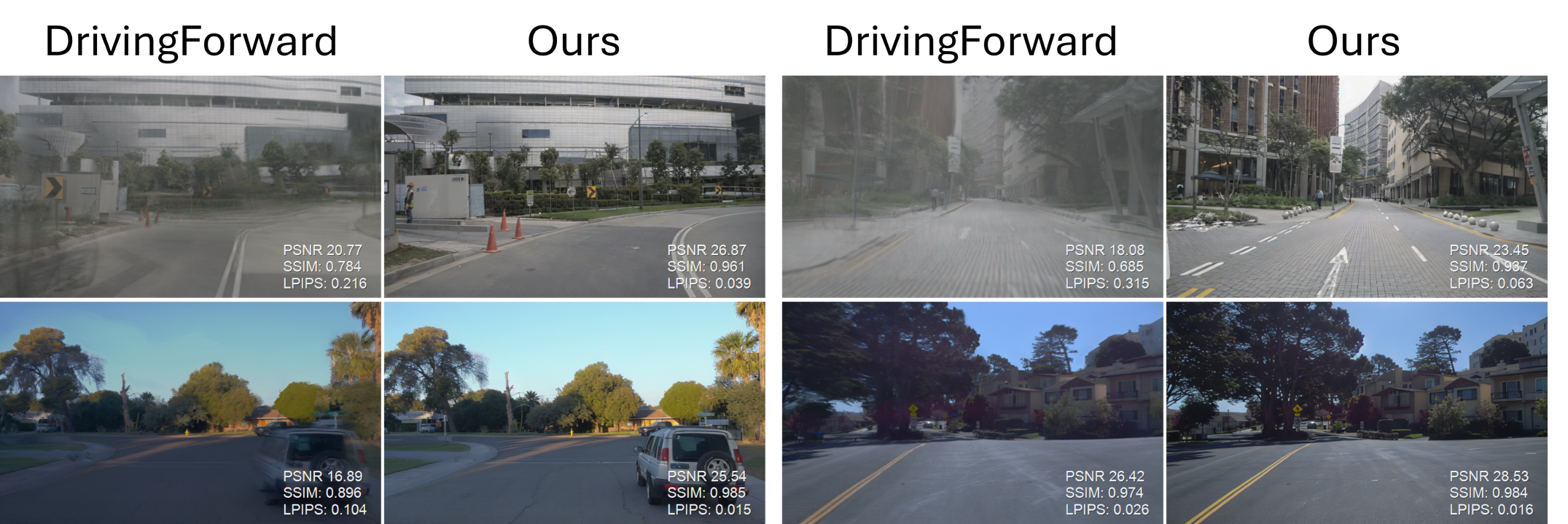}
\caption{\textbf{Qualitative comparison of DrivingForward}~\cite{tian2025drivingforward} before and after \myMethod enhancement.}
    \label{fig:fix_comp_drivfwd}
\end{figure}

\begin{figure}[t!]
   \centering
\includegraphics[width=0.88\textwidth]{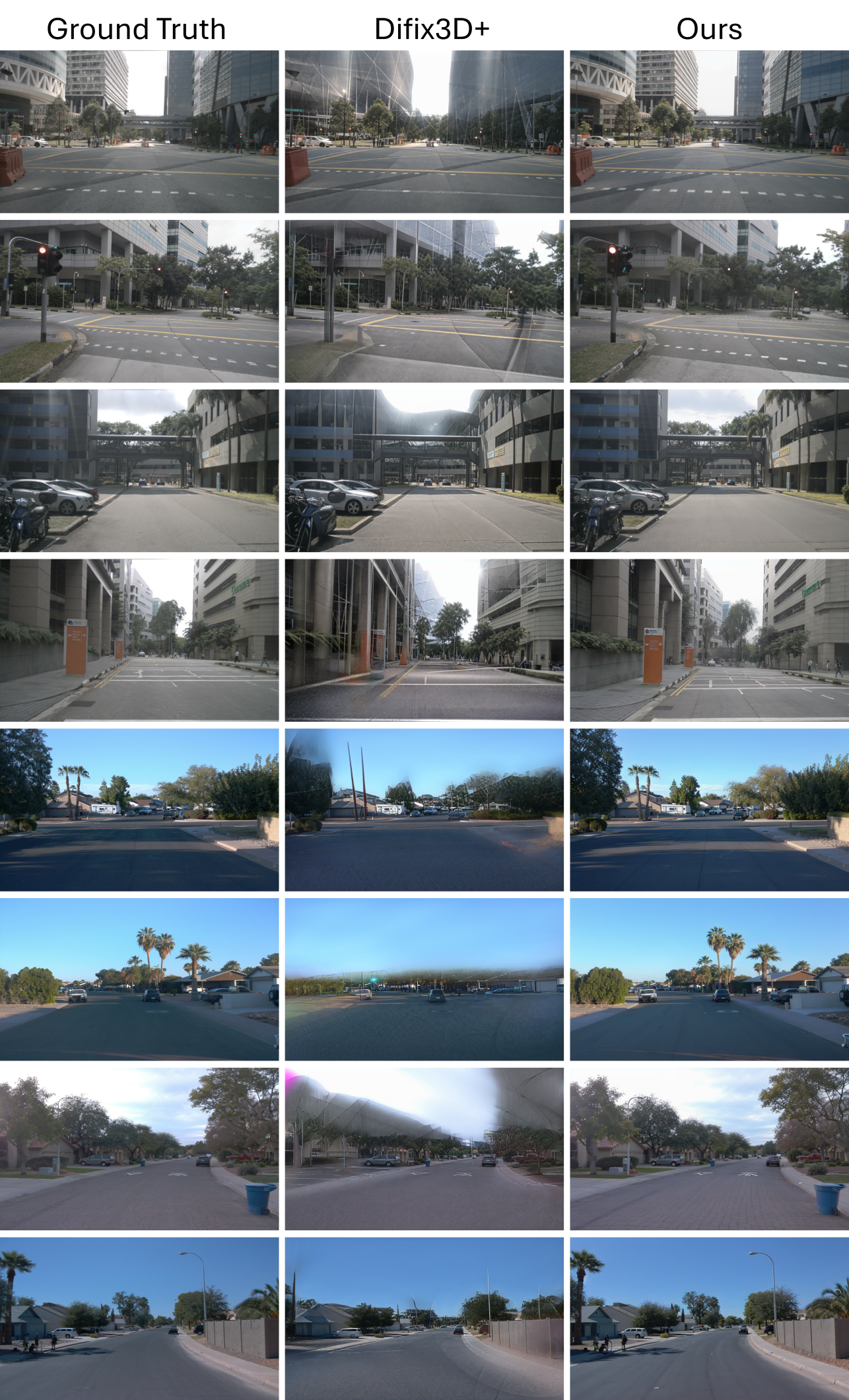}
   \caption{\textbf{Qualitative comparison with Difix3D+}~\cite{wu2025difix3d+}. We use DepthSplat~\cite{xu2025depthsplat} as the feedforward backbone for our approach.}
    \label{fig:vis_comp_difix3D+}
\end{figure}

\subsection{Novel View Synthesis on New Trajectories}
\label{sec:nvs}
Beyond the original camera trajectory, we additionally render novel views along \emph{out-of-distribution} viewpoints in Fig.~\ref{fig:ood_traj}, including lateral offsets (\eg, 1\,m and 3\,m) and a virtual drone-like camera (about 2.5\,m height with a 20$^\circ$ downward pitch). Qualitatively, \myMethod remains stable under these more challenging trajectory shifts, producing consistent geometry and appearance without noticeable artifacts. This trajectory-level controllability suggests a promising direction for data augmentation in driving-centric novel view synthesis benchmarks, enabling new viewpoints that are absent from the original dataset. In particular, it opens up opportunities for (\emph{i}) introducing drone viewpoints for driving scenes and (\emph{ii}) synthesizing multi-agent perspectives for collaborative perception~\cite{yu2022dair,xu2023v2v4real,song2024collaborative,zhou2024v2xpnp,zimmer2024tumtraf,zhou2025turbotrain}. 

\subsection{Plug-and-Play Enhancement}
\label{sec:further_backbone}

We further illustrate our enhancement on DepthSplat in Figs.~\ref{fig:vis_comp_depthsplat-1} and \ref{fig:vis_comp_depthsplat-2} and on WorldMirror in Fig.~\ref{fig:fix_comp_hunyuan}.
Moreover, to demonstrate that our approach is plug-and-play, we apply \myMethod as an enhancement module on top of additional feedforward backbones, including AnySplat~\cite{jiang2025anysplat} and DrivingForward~\cite{tian2025drivingforward}.
Figs.~\ref{fix_comp_anysplat} and \ref{fig:fix_comp_drivfwd} report the corresponding before and after comparisons for AnySplat and DrivingForward, respectively.
Using the same experimental settings, we conduct qualitative comparisons, showing consistent improvements across diverse backbones and highlighting the generality of our method. 

\subsection{Visual Comparison with Difix3D+}
\label{sec:difix3D}

We further compare against Difix3D+~\cite{wu2025difix3d+} to examine the iterative 3D optimization and diffusion fixing paradigm under sparse-view driving scenarios. Our reproduced Difix3D+ is built on gsplat 3DGS trained from scratch for 30k steps; every 3k steps, we apply Difix to all novel view renderings and replay the refined outputs as additional supervision (10 rounds total). We set the novel view loss weight to $0.3$ and the SSIM weight to $0.2$. For diffusion inference, we use the official implementation with all default settings.

As shown in Fig.~\ref{fig:vis_comp_difix3D+}, Difix3D+ typically produces brighter, sharper textures and visually crisper renderings, but it also tends to introduce prominent ghosting patterns, while distant regions are often over-smoothed or blurred. We conjecture this stems from the tight coupling between diffusion fixing and the 3D optimization loop: once diffusion-refined signals are injected during training, intermediate Gaussians can be prematurely driven by diffusion-induced residuals before the geometry is sufficiently constrained by multi-view evidence. Applying diffusion iteratively further compounds this effect, accumulating style shifts across rounds and effectively amplifying errors rather than correcting them, which manifests as repeated textures, novel view inconsistency, and degraded far-range fidelity. In contrast, our method derives a per-pixel confidence map via reprojection-based cross-validation and uses it to gate both modulated densification and gradient updates. This effectively suppresses inconsistent diffusion residuals from being absorbed into 3D, yielding more stable structures and improved cross-view consistency while retaining visual enhancement.

\clearpage